\documentclass[authoryear,10pt]{article}

\usepackage{amssymb,multirow,amsmath,graphicx,array,lscape,rotating}
\usepackage{geometry}
\usepackage{graphicx}
\usepackage{subcaption}

\usepackage{PRIMEarxiv}
\usepackage{algpseudocode}

\usepackage[utf8]{inputenc} 
\usepackage[T1]{fontenc}    
\usepackage{hyperref}       
\usepackage{url}            
\usepackage{booktabs}       
\usepackage{amsfonts}       
\usepackage{nicefrac}       
\usepackage{microtype}      
\usepackage{lipsum}
\usepackage{fancyhdr}       
\usepackage{graphicx}       
\graphicspath{{media/}}     
\usepackage{amssymb}
\usepackage{amsthm}
\usepackage{amsmath}
\usepackage{tabularx}
\usepackage{booktabs}
\usepackage{multicol}
\usepackage{relsize}
\usepackage{float}
\usepackage{xurl}
\usepackage{hyperref}
\usepackage{graphicx}
\usepackage{tablefootnote}

\usepackage{setspace}
\usepackage{xcolor}

\title{Improving the Forecast Accuracy of Wind Power by Leveraging Multiple Hierarchical Structure}
\author{
  Lucas English\\
  School of Mathematics and Physics\\
  The University of Queensland\\
  St Lucia, QLD 4072, Australia\\
  \texttt{lucas.english@uq.net.au} \\
   \And
  Mahdi Abolghasemi\\
  School of Mathematics and Physics\\
  The University of Queensland\\
  St Lucia, QLD 4072, Australia\\
  \texttt{m.abolghasemi@uq.edu.au} \\
}

\begin{document}
\maketitle

\begin{abstract}
Renewable energy generation is of utmost importance for global decarbonization. Forecasting renewable energies, particularly wind \textcolor{black}{energy}, is challenging due to the inherent uncertainty in wind \textcolor{black}{energy} generation, which depends on weather conditions. Recent advances in hierarchical forecasting through reconciliation have demonstrated a significant increase in the quality of wind \textcolor{black}{energy} forecasts for short-term periods. We leverage the cross-sectional and temporal hierarchical structure of turbines in wind farms and build cross-temporal hierarchies to further investigate how integrated cross-sectional and temporal dimensions can add value to forecast accuracy in wind farms.  We found that cross-temporal reconciliation was superior to individual cross-sectional reconciliation at multiple temporal aggregations. Additionally, \textcolor{black}{machine learning based forecasts that were cross-temporally reconciled demonstrated high accuracy at coarser temporal granularities,} which may encourage adoption for short-term wind forecasts. Empirically, we provide insights for decision-makers on the best methods for \textcolor{black}{forecasting high-frequency wind data across different forecasting horizons and levels}.
\end{abstract}

\keywords{Hierarchical forecasting reconciliation \and Renewable energy \and Wind \textcolor{black}{energy} forecasting \and Cross-sectional forecasts \and Cross-temporal forecasts}

\section{Introduction}
\label{sec:sample1}
Modern forecasting relies on the axiom that not only the history of one time series can be utilized in predicting future events but the histories of other series can be useful also. For time series forecasting, patterns in historical measurements are used to extrapolate future values of the series. \textcolor{black}{In many circumstances, time series can be aggregated to form hierarchical frameworks \cite{HYNDMAN20112579}}. Aggregation may be performed logically, geographically, temporally, etc. \textcolor{black}{While there are advantages and disadvantages associated with each of these approaches, it has been shown that building time-series hierarchies can improve forecast accuracies across locations and various forecasting horizons in different problems \cite{Athanasopoulos2020}, including renewable energies sectors \cite{DIFONZO202313}.} For example, in forecasting wind \textcolor{black}{energy}, we can either forecast at the total aggregated level which may miss information from turbine locations and their individual performance, or we can forecast each turbine individually considering all of their unique \textcolor{black}{information \cite{https://doi.org/10.1002/we.2819}. Such hierarchical wind energy forecasts can be produced by aggregating data into cross-sectional and temporal hierarchies.} Ensuring the coherency of forecasts of the hierarchy can be achieved through different reconciliation algorithms, however, there is not enough empirical evidence to depict how reconciliation methods can help to improve the forecasting performance across different levels and horizons, and whether hierarchical forecasting can be useful in a wind-farm setting for very short-term forecasting.

Renewable energy technologies are considered to be a key challenge for facing climate change and energy security problems \textcolor{black}{\cite{LeeChul-Yong2017Ftdo}}. Forecasting renewable energies is of utmost importance for the security of the systems and ensuring consistent supply for demand. One leading source of renewable energy is wind energy, which contributed \textcolor{black}{7.3\% of the global electricity demand in 2022, and is projected to account for 12.1\% by 2028 \cite{iea}.} \textcolor{black}{However, accurate forecasting of wind energy is non-trivial given the inherently variable and uncertain nature of wind speeds on different time scales}. Accurate forecasting of wind energy generation is critical for planning suitable \textcolor{black}{energy} schedules and setting achievable renewable energy targets, on the scale of minutes to days-ahead \cite{abolghasemi2021state, petropoulos2022forecasting}. Hierarchical forecasting ensures coherency of forecasts, and has been empirically shown to improve wind energy forecasts \cite{8453006, sharma2023optimal}. However, difficulties involved in forecasting wind \textcolor{black}{energy} includes the limited predictability of the weather, failure of turbines due to technical or physical issues such as high wind speed, and the nonlinearity of the power curve of wind turbines \cite{NorbuSonam2021Mtro}. Recent studies leverage the cross-sectional hierarchical time series and show that forecast accuracy can be improved by reconciliation methods \cite{sharma2023optimal}. These methods are suitable for determining optimal forecasts for wind turbines and wind farms located in different geographical locations. However, forecasts are often required for various forecasting horizons, including very short-term to short-term and longer terms. The temporal information in data can be used for building temporal hierarchies and cross-temporal hierarchies to further improve forecast accuracy across different levels and horizons. 

In this study, we sought to empirically identify the best strategy for reconciling the hierarchical time series in the cross-sectional and cross-temporal paradigms for \textcolor{black}{high-frequency} wind energy data. We aim to improve the forecast accuracy for wind energy in the very short-term horizon, i.e., from 10 minutes up to one hour, across different levels of the hierarchy. To shed light on the best approaches for short-term forecasting on various horizons, we investigate whether it is advantageous to first temporally aggregate historical data, as needed for decision making, and use cross-sectional methods on the temporally aggregated data for forecasting across different levels, or \textcolor{black}{tap into the more sophisticated cross-temporal techniques to generate coherent forecasts across both temporal and cross-sectional dimensions}. We compare accuracies through 10-minute, 20-minute, 30-minute, and one-hour-ahead forecasts to mimic practical forecasting concerns. We compare a variety of cross-sectional and cross-temporal forecasting reconciliation techniques to identify the best setting using Friedman and post-hoc Nemenyi tests. We use two models for generating base forecasts, including a simple linear regression and \textcolor{black}{Light gradient boosting machine} as a popular powerful machine learning (ML) model that has shown promising performance in the M5 hierarchical forecasting competition \cite{MAKRIDAKIS20221346}. \textcolor{black}{Our contribution to the literature is as follows}: 
\begin{itemize}
    \item \textcolor{black}{We determine the value of hierarchical time series forecasting methods on forecasting accuracy of wind energy for various short-term horizons \textcolor{black}{using one-step ahead and direct multi-step ahead forecasting models} (10, 20, 30 minutes, and one hour).}
    
    \item \textcolor{black}{We investigate the effectiveness of temporal aggregation followed by cross-sectional forecasting methods against the direct application of cross-temporal forecasting to shed light on the value of hierarchical time series reconciliation techniques for high-frequency data of wind farms.}

    \item \textcolor{black}{We empirically evaluate the accuracy of simple regression models and more sophisticated gradient-boosting machines - that have shown promising performance for low-frequency data in hierarchical time series - for high-frequency wind energy data.}
\end{itemize}

The rest of this article is organised as follows. Section \ref{sec:background} reviews the background and literature of wind \textcolor{black}{energy} forecasting. Section \ref{sec:methods} outlines the most common hierarchical forecasting methods and discusses their advantages and disadvantages. \textcolor{black}{Section \ref{sec:base} introduces the two base forecasting methods used in this work, namely linear regression and LightGBM.} Section \ref{sec:data} presents the primary data sets used in this work, describes the preprocessing method, and explains the experimental setup. Section \ref{sec:results} presents the empirical results of the experiment and discusses our findings. Finally, Section \ref{sec:conclusion} concludes the paper.

\section{Background}
\label{sec:background}
The study of wind \textcolor{black}{energy} forecasting has traditionally centered around improving \textcolor{black}{forecast accuracy of models for individual time series, with much less attention to hierarchical information in data and using hierarchical reconciliation methods to improve their accuracy.} Wind \textcolor{black}{energy} base forecasts can be classified by utilizing physical \textcolor{black}{\cite{lange2006physical}}, statistical \cite{9435209}, deep learning methods \cite{deep}, Bayesian modeling \cite{jiang2013very}, or hybrids thereof \cite{HOSSAIN2021126564}. Physical methods rely on modeling the atmospheric conditions and topographical features that affect wind patterns. This may include numerical weather predictions (NWP) which can rely on historical meteorological data such as wind speed, direction, temperature and pressure. Additionally, physical models often use the power curve of the turbines of interest to extrapolate wind predictions to \textcolor{black}{energy} forecasts. Physical methods excel in medium and long-term forecast horizons \cite{en13153764}, however \textcolor{black}{energy} forecast accuracy degrades with that of weather predictions \cite{app112110335}. Statistical methods rely on developing time series models of wind \textcolor{black}{energy} from historical data, typically without considering meteorological conditions \cite{WANG2011770}. Statistical methods demonstrate high accuracy in the very short and short-term horizon, from several minutes to several hours-ahead \cite{en13153764}. Some common techniques include \textcolor{black}{linear regression \cite{DEMOLLI2019111823} autoregressive moving average (ARMA) \cite{LYDIA2016115} and gradient boosting \cite{en16031132} models.} Deep learning methods utilize multilayer neural networks to extract the features of wind \textcolor{black}{energy} time series data to predict the output using classification results. Deep learning methods include support vector machines (SVM) \cite{LIU2020106056}, Kalman filters \cite{ALY2022122367} and recurrent neural networks (RNN) \cite{deep}, and have demonstrated improved forecasting accuracy over statistical methods in many previous studies \cite{WANG2021117766}.

Very short-term forecasts, including minutes to a few hours ahead, are important for decision-making of wind \textcolor{black}{energy} generation and for such forecasts, recent observations are more relevant to forecasting than NWP \textcolor{black}{\cite{SweeneyConor2020Tfof}}. The volatility of wind \textcolor{black}{energy} generation necessitates such short-term forecasts to allow for wind farm operators to enable backup power supplies in case of shortages \cite{BlonbouRuddy2011Vswp}. Common forecast horizons in literature include 10-minutely, 20-minutely, 30-minutely, and 1-hourly \cite{petropoulos2022forecasting, BlonbouRuddy2011Vswp} and accurate forecasts at each horizon can be important for power distribution planning. Strong complex seasonality in the form of multiple seasonality has been observed in instantaneous wind speeds \cite{azorin2021decline}, and thus must be considered with extra care in forecasting. This is, in particular, challenging in the context of hierarchical forecasting because temporal aggregation may change the dynamic of multiple seasonality in data and introduce a new one. The optimal hierarchical forecasting method is closely related to the characteristics of the individual series involved, forecasting horizon, the level at which forecasts are needed, among other factors \cite{abolghasemi2022model}. \textcolor{black}{In achieving forecasts that are coherent across both space and time, one could use cross-temporal hierachies and their related simultaneous reconciliation techniques, or alternatively build various cross-sectional hierarchical time series models for the required temporal frequencies and generate forecasts for these hierarchies to obtain coherent forecasts across different levels for individual temporal frequencies. While building integrated cross-temporal hierarchies will take more computational effort and expertise, building cross-sectional hierarchies is easier.  It is not evident if we can obtain significantly more accurate forecasts using the cross-temporal hierarchies for high-frequency wind data that justify the additional computational and technical effort.}

Temporal aggregation transforms a time series of a given frequency to a series of a lower frequency. For example, a time series of length $n$ with hourly measurements may be temporally aggregated into a new series of length $n/2$ with 2-hourly measurements, or length $n/24$ with daily measurements, among other possibilities. For fast-moving series, higher level aggregations capture the trend patterns better since the higher frequency components of the underlying series are averaged out. However, lower aggregation levels are better for investigating seasonal patterns \cite{kourentzes2014improving}.

\textcolor{black}{Improving forecasting accuracy through temporal aggregation of data originates in economics literature \cite{AmemiyaTakeshi1972TEoA}. However, in 2014, multiple temporal aggregation (MTA) was first proposed, whereby aggregation at multiple frequencies was performed to extract useful information at varying levels of temporal granularity \cite{kourentzes2014improving}. The multiple aggregation prediction algorithm (MAPA) consists of three steps. Firstly, data from a base-level time series is aggregated temporally into non-overlapping series. For example, if we have an aggregation level of 2, then the mean of every two observations is taken as an element without overlaps, creating a new time series of length $n/2$. Secondly, base forecasts are performed on each time series, conventionally using exponential smoothing methods, at each temporal aggregation. Thirdly, the parameters of the final model are extracted by combining the parameters of the base models. For example, the error and trend parameters can be set as the unweighted mean of base forecast models, and care must be taken to only average seasonal components when aggregation levels are consistent with seasonality \cite{kourentzes2014improving}. Disaggregated temporal time series must sum up to the higher levels, similar to cross-sectional hierarchies, to ensure coherency. Forecasting with temporal aggregation has been shown empirically to produce more accurate forecasts even if they are theoretically suboptimal \cite{kourentzes2017demand}.}

\textcolor{black}{Similar to logical and geographic aggregation, temporal aggregation can also be considered under the hierarchical time series framework. In 2017, temporal hierarchical forecasting (THF) was proposed, which similarly utilises temporally aggregated data with hierarchical reconciliation techniques \cite{AthanasopoulosGeorge2017Fwth}. By indexing all time series by the index of the highest-level aggregated time series, one can construct a hierarchy. In these hierarchies, for each observation of the highest-aggregated level, there will be progressively more observations at lower-aggregated levels. Compared to MAPA, THF allows one to leverage the wealth of hierarchical reconciliation techniques from literature directly, which has demonstrated increased forecasting accuracy empirically \cite{AthanasopoulosGeorge2017Fwth,NystrupPeter2021Drif}. Issues arise with estimating the h-step-ahead base forecast error covariance $\mathbf{W}_{h}$, as the time series at the aggregated levels will have the number of samples reduced by a factor of the respective aggregation levels. New estimates of $\mathbf{W}_{h}$ based on autocorrelation of forecast errors \cite{NystrupPeter2020Thwa}, and eigendecomposition of the temporal correlation matrix \cite{NystrupPeter2021Drif} have been proposed to overcome this problem.}

An emerging method in hierarchical forecasting is cross-temporal forecasting. In cross-temporal hierarchies, aggregation is performed across both temporal as well as a logical or geographic scale. \textcolor{black}{That is, there is a cross-sectional hierarchy as well as a temporal hierarchy. Cross-temporal hierarchies were first proposed in 2019 \cite{kourentzes2019cross}. It was shown that by simply performing cross-sectional, then temporal reconciliation sequentially (or vice versa), coherency across the cross-temporal hierarchy is not guaranteed. In 2023, an iterative cross-temporal reconciliation method was proposed in which cross-sectional and temporal reconciliation was performed iteratively until sufficiently small incoherence was achieved \cite{DiFonzoTommaso2023CfrO}. Also in 2023, the theoretically optimal combination simultaneous reconciliation was proposed \cite{DiFonzoTommaso2023CfrO}, which follows from the work of \cite{HYNDMAN20112579,wickramasuriya2019optimal} applied to the full cross-temporal hierarchy. It has been empirically suggested in literature that cross-temporally reconciled forecasts may not offer substantial accuracy gains over reconciliation only across the temporal dimension \cite{ATHANASOPOULOS2023}. However, for decision-making, it may be necessary to have cross-temporally coherent forecasts \cite{kourentzes2019cross}.} In achieving forecasts that are coherent across both space and time, one could use cross-temporal hierarchies and their related simultaneous reconciliation techniques, or \textcolor{black}{ alternatively build various cross-sectional hierarchical time series models for the required temporal frequencies and generate forecasts for these hierarchies to obtain coherent forecasts across different levels for individual temporal frequencies. While building integrated cross-temporal hierarchies will take more computational effort and expertise, building cross-sectional hierarchies is easier.  It is not evident if we can obtain significantly more accurate forecasts using the cross-temporal hierarchies for high-frequency wind data that justify the additional effort.}

\textcolor{black}{Simultaneous reconciliation across the cross-sectional and temporal hierarchies is a non-trivial problem, and it remains as a challenging problem for large hierarchies. One reason for this difficulty is that the cross-temporal hierarchy is represented by the tensor product across the temporal and cross-sectional dimensions. Therefore, the number of nodes of the cross-temporal hierarchy is equal to the product of the number of nodes in the cross-sectional hierarchy, and the temporal hierarchy. As a result, the hierarchy can become prohibitively large \cite{kourentzes2019cross}. However, recent work into cross-temporal forecast reconciliation has largely overcome these difficulties through computationally-feasible estimates of $\mathbf{W}_{h}$ \cite{GIROLIMETTO2023,DiFonzoTommaso2023CfrO}. One focal point of current literature is probabilistic cross-temporal reconciliation, in which individual forecasts are given as probability distributions which lie in the \textit{coherent probability space} defined in \cite{GIROLIMETTO2023}. However, probabilistic forecasting lies outside of the scope of this work and will not be considered. Nonetheless, deterministic cross-temporal reconciliation has empirically shown to improve forecasting accuracy in a variety of forecasting scenarios \cite{kourentzes2019cross, DiFonzoTommaso2023CfrO,DIFONZO202313}.}

\section{Hierarchical forecasting methods}\label{sec:methods}
Reconciliation towards coherent forecasts of hierarchical time series can be largely broken down into four categories: bottom-up (BU), top-down (TD), middle-out (MO) and combination (COM). In the BU paradigm, base forecasts are produced at the bottom level of the hierarchy, and are summed to give forecasts for higher level nodes in the hierarchy. In the TD paradigm, base forecasts are performed on the root node, and the historical proportionality of the lower-level nodes are used to disaggregate the forecast \cite{gross1990disaggregation}. The MO method performs base forecasts at an intermediate level. Then, the forecasts are summed upwards on the hierarchy, and historical proportions are used to disaggregate forecasts to lower-level nodes. Forecasts of the BU, TD and MO paradigms are inherently coherent, that is, forecasts at every level sum to give values which are consistent with the aggregation structure.

The COM method produces forecasts at multiple levels of the hierarchy and \textcolor{black}{taps into the information across all levels to} perform reconciliation \cite{wickramasuriya2019optimal}. One distinguished method within the COM paradigm is the trace minimization (MinT) algorithm. MinT has a strong theoretical foundation \cite{wickramasuriya2019optimal}, and has empirically shown to improve forecast accuracies in \textcolor{black}{several studies} \cite{wickramasuriya2019optimal, ATHANASOPOULOS2023, DIFONZO202313}. However, it has been suggested that MinT may provide inadequate forecasts when in-sample errors do not represent post-sample accuracy of the baseline forecast models, or when the hierarchy of interest places more emphasis on some levels than others \cite{abolghasemi2022machine}.

\begin{figure}[H]
\centering
\includegraphics[scale=0.7]{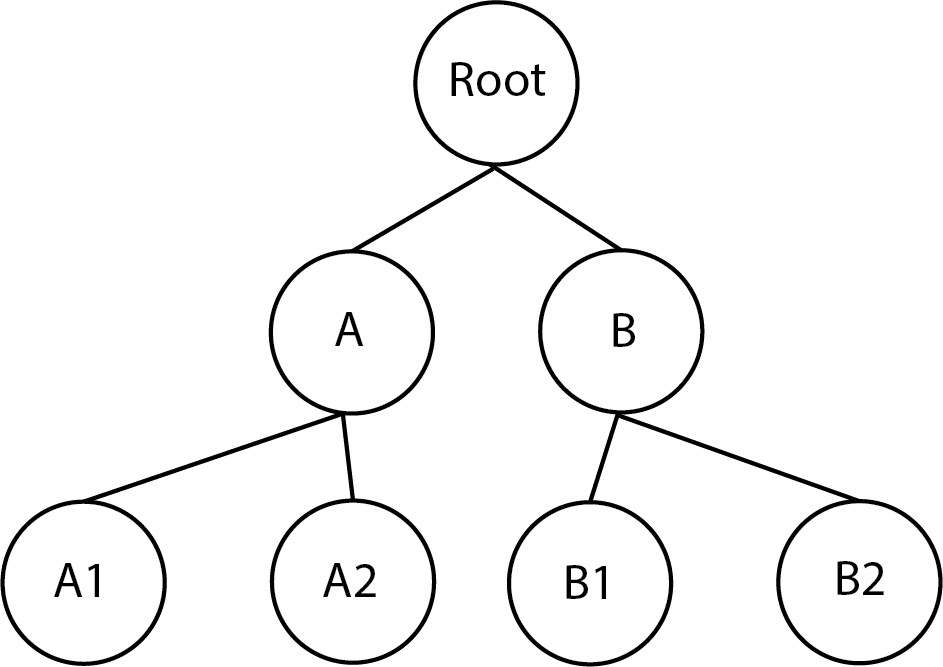}
\caption{Example time series hierarchy}\label{fig:hierarchy}
\end{figure}

We provide herein a basic overview of the methods, and the notations and parameters used in this study. \textcolor{black}{We adopt the notations proposed by \cite{hyndman2022}. Let a hierarchical time series be a multivariate time series denoted by $\mathbf{y}_{1},\cdots,\mathbf{y}_{T}$ satisfying linear constraints that reflect some known structure. An example time series hierarchy is given in Figure \ref{fig:hierarchy}. To such a hierarchy, we define the following parameters, in which bold variables signify vectors or matrices:}\\

\color{black}
\makebox[1.5cm]{$n:$}  Total number of nodes (i.e., time series) in the hierarchy.\par
\makebox[1.5cm]{$n_{i}:$}  Total number of nodes at depth $i$, e.g., $n_{0}=1$.\par
\makebox[1.5cm]{$k:$}  The tree height of the hierarchy, with $0\leq i \leq k$.\par
\makebox[1.5cm]{$n_{b}:$}  The total number of disaggregated series ($n_{b}=n_{k}$).\par
\makebox[1.5cm]{$n_{a}:$} The total number of aggregated series, with $n_{a}=n-n_{b}$.\par
\makebox[1.5cm]{$\mathbf{b}_{t}:$} A vector of $n_{b}$ bottom-level time series at time $t$.\par
\makebox[1.5cm]{$\mathbf{a}_{t}:$} A vector of $n_{a}$ aggregated time series at time $t$.\par
\makebox[1.5cm]{$\mathbf{y}_{t}:$} A vector holding the time series of all nodes at time $t$, such that $ \mathbf{y}_{t}=\left[\begin{matrix}\mathbf{a}_{t} \\ \mathbf{b}_{t}\end{matrix}\right] $.\par
\makebox[1.5cm]{$\mathbf{S}:$} The \textit{summing} or \textit{structural} matrix that encodes the linear constraints, such that $\mathbf{y}_{t}=\mathbf{S}\mathbf{b}_{t}$. We can write this matrix as $S=\left[\begin{matrix}\mathbf{A} \\ \mathbf{I}_{n_{b}}\end{matrix}\right]$, where $\mathbf{A}$ is the aggregation matrix that encodes how the bottom level series aggregate to higher levels, and $\mathbf{I}_{n_{b}}$ is the $n_{b} \times n_{b}$ identity matrix. The columns of $\mathbf{S}$ form a linear subspace of $\mathbb{R}^{n}$ in which the linear constraints hold. We call this the \textit{coherent subspace}, and denote it as $\mathfrak{s}$.\par
\makebox[1.5cm]{$\mathbf{\hat{y}}_{h}:$} The h step-ahead base forecasts of $y_{T+h}$ for all nodes from $T$ historical values.\par
\makebox[1.5cm]{$\mathbf{\tilde{y}}_{h}:$} The h step-ahead forecasts of $y_{T+h}$ for all nodes from $T$ historical values after reconciliation has projected the original vector onto $\mathfrak{s}$ using a chosen reconciliation method.\par

For example, using the notation above, the hierarchical time series in Figure \ref{fig:hierarchy} can be expressed as

\begin{align}
\mathbf{y}_{t}&=\mathbf{S}\mathbf{b}_{t},\nonumber\\
\begin{bmatrix}
y_{t}\\
y_{A,t}\\
y_{B,t}\\
y_{A1,t}\\
y_{A2,t}\\
y_{B1,t}\\
y_{B2,t}\\
\end{bmatrix}&=
\begin{bmatrix}
1 & 1 & 1 & 1\\
1 & 1 & 0 & 0 \\
0 & 0 & 1 & 1\\
 &  & I_{4} & 
\end{bmatrix}
\color{black}\begin{bmatrix}
y_{A1,t}\\
y_{A2,t}\\
y_{B1,t}\\
y_{B2,t}
\end{bmatrix}.\nonumber
\end{align}

With the knowledge of the hierarchical structure, we can determine $\mathbf{S}$, and various reconciliation methods can be expressed through $\mathbf{\tilde{y}}_{h}=\mathbf{S}\mathbf{G}\mathbf{\hat{y}}_{h}$, for some matrix $\mathbf{G}$. The matrix $\mathbf{G}$ is of dimension $n_{b}\times n$ and is determined by the reconciliation method used, which determines how we project the incoherent base forecasts $\mathbf{\hat{y}}\in\mathfrak{s}^{\perp}$ onto the coherent subspace $\mathfrak{s}$. For example, with the hierarchy in Figure \ref{fig:hierarchy}, $\mathbf{G}$ for BU reconciliation is expressed as

\begin{equation}
    \mathbf{G}=\begin{bmatrix}
    0 & 0 & 0 & 1 & 0 & 0 & 0\\
    0 & 0 & 0 & 0 & 1 & 0 & 0\\
    0 & 0 & 0 & 0 & 0 & 1 & 0\\
    0 & 0 & 0 & 0 & 0 & 0 & 1\\
    \end{bmatrix}=[\mathbf{0}_{n_{b}\times n_{a}}\vert \mathbf{I}_{n_{b}}].\nonumber
\end{equation}

For TD reconciliation, $\mathbf{G}$ is expressed as

\begin{equation}
    \mathbf{G}=\begin{bmatrix}
    p_{1} & 0 & 0 & 0 & 0 & 0 & 0\\
    p_{2} & 0 & 0 & 0 & 0 & 0 & 0\\
    p_{3} & 0 & 0 & 0 & 0 & 0 & 0\\
    p_{4} & 0 & 0 & 0 & 0 & 0 & 0\\
    \end{bmatrix}=[\mathbf{g}\vert \mathbf{0}_{n_{b}\times(n-1)}].\nonumber
\end{equation}
\color{black}

In this expression, $\mathbf{g}=[p_{1},p_{2},\cdots p_{k}]$. The values $p_{1}$, $p_{2}$, $p_{3}$ and $p_{4}$ represent historical proportions which distribute the base forecast of the root node down to the bottom level. \textcolor{black}{There are various methods to compute these proportions, including the \textit{average historical proportions} and the \textit{proportions of historical averages}, but we opt for \textit{forecast proportions}, proposed in \cite{ATHANASOPOULOS2009146} as it is able to capture changing proportions in the hierarchy over time. In this method, the proportions of the forecasts at each step are used to propagate down the base forecast from the top-level. For a $k$-level hierarchy, the proportions are computed through}
\textcolor{black}{
\begin{equation}
    p_{j}=\prod_{l=0}^{K-1}\frac{\hat{y}^{(l)}_{j,h}}{\hat{S}_{j,h}^{(l+1)}},
\end{equation}}

\textcolor{black}{where $j=1,2,\cdots,n_{b}$, $\hat{y}^{(l)}_{j,h}$ is the h-step-ahead forecast of the series for the node $l$ levels above node $j$ and $\hat{S}_{j,h}^{(l+1)}$ is the sum of the h-step-ahead base forecasts that are $l$ levels above node $j$ and are directly connected to the node that is $l+1$ levels above node $j$. For example, in the hierarchy in Figure \ref{fig:hierarchy}, we can compute}

\textcolor{black}{\begin{equation}
    p_{A1}=(\frac{\hat{y}_{A1,h}}{\hat{y}_{A1,h}+\hat{y}_{A2,h}})(\frac{\hat{y}_{A,h}}{\hat{y}_{A,h}+\hat{y}_{B,h}}).\nonumber
\end{equation}}

MO methods will have $\mathbf{G}$ matrices \textcolor{black}{with a form that resembles a combination of the BU and TD methods. The precise form will depend on the level of the hierarchy at which MO reconciliation is performed.}

The COM method produces forecasts across all nodes of the hierarchy and combines them linearly to reconcile the forecasts to ensure coherency. The covariance matrix of the errors of forecasts is given by\footnote{A proof for this equality can be found in \cite{wickramasuriya2019optimal}.}

\begin{equation}
    \mathbf{V_{h}} = \text{Var}[\mathbf{y}_{n+h}-\mathbf{\tilde{y}}_{n+h}]=\mathbf{S}\mathbf{G}\mathbf{W}_{h}\mathbf{G}'\mathbf{S}',
\end{equation}

\textcolor{black}{where} the matrix $\mathbf{W}_{h}$ is the variance-covariance matrix of the $h$-step-ahead base forecast errors from the forecasts at every node. The diagonal elements of $\mathbf{V}_{h}$ represent the autocovariance of the errors. Therefore, to produce unbiased reconciled forecasts, it is desireable to minimize the trace of $\mathbf{V}_{h}$. The matrix $\mathbf{G}$ which minimizes the trace of $\mathbf{V}_{h}$ is

\begin{equation}\label{eq:G_matrix}
    \mathbf{G}=(\mathbf{S}'\mathbf{W}_{h}^{\dagger}S)^{-1}\mathbf{S}'\mathbf{W}_{h}^{\dagger},
\end{equation}

\textcolor{black}{where} $\mathbf{W}_{h}^{\dagger}$ is the generalized inverse of $\mathbf{W}_{h}$. Evaluating $\mathbf{W}_{h}$ can be computationally expensive, but there are several simplifying approximations, including the methods of ordinary least squares (OLS), variance scaling, structural scaling and the shrinkage estimator \cite{hyndman2018forecasting}. In this work, OLS and the shrinkage estimator will be considered for their low time-complexity and efficiency in dealing with large hierarchies \cite{SPILIOTIS2021107756}.

For the OLS method, $\mathbf{W}_{h}$ is set to the identity matrix, $\mathbf{W}_{h}=\mathbf{I}$. One disadvantage is that this specification does not distinguish between differences in scale of the hierarchy. 

\color{black}
For the shrinkage estimator, we must first define the unbiased sample estimator of the in-sample one-step-ahead base forecast errors as

\begin{equation}
    \mathbf{\hat{W_{1}}}=\frac{1}{T}\sum_{t=1}^{T}\mathbf{\hat{e}}_{t}(1)\mathbf{\hat{e}}_{t}(1)'.
\end{equation}

Note this is simply the covariance matrix of the residuals for the in-sample one-step-ahead forecasts. One can easily derive it using the usual formula for covariance, noting all time series must have zero mean for residuals across samples to be unbiased. The prefactor $1/T$ is used rather than $1/(T-1)$ for unbiased sample covariance estimation with Bessel's correction as we assume the population mean is known and equal to zero. The in-sample h-steps-ahead base forecast errors are given by

\begin{equation}
    \mathbf{\hat{e}}_{t}(h)=\mathbf{y}_{t+h}-\mathbf{\hat{y}}_{t+h}.
\end{equation}

The shrinkage estimate of $\mathbf{W}_{h}$ is then computed through

\begin{equation}
    \mathbf{W}_{h}=\lambda_{D}\mathbf{\hat{W}}_{1,D}+(1-\lambda_{D})\mathbf{\hat{W}}_{1},
\end{equation}

\textcolor{black}{where} $\mathbf{\hat{W}}_{1,D}$ is a diagonal matrix with diagonal entries of $\mathbf{\hat{W}}_{1}$, and $\lambda_{D}$ is the shrinkage parameter \cite{schafer2005shrinkage}. Off-diagonal elements of $\mathbf{\hat{W}}_{1}$ are shrunk towards zero (hence shrinkage), while diagonal elements, i.e., variances, are unchanged. To compute the shrinkage parameter, we first standardize our residuals, such that they have unit variance (with the mean already being 0). That is,

\begin{equation}
    z_{i,t} = \frac{e_{i,t}}{\sqrt{\frac{1}{T}\sum_{t=1}^{T}(e_{i,t})^{2}}}.
\end{equation}

Using the standardized residuals, we compute the 1-step-ahead in-sample correlation matrix through\footnote{\textcolor{black}{Note the covariance of standardized variables gives the correlation under the normal distribution assumption \cite{KimHae-Young2018Snfc}. This is used computationally in \cite{foreco}.}}

\begin{equation}
    \overline{r}_{ij}=\frac{1}{T}\sum_{t=1}^{T}z_{i,t}z_{j,t}.
\end{equation}

The variance of the elements of this correlation matrix is computed through

\begin{align}
    \text{Var}(\overline{r}_{ij})&=\frac{1}{T}\text{Var}(r_{ij})=\frac{1}{T(T-1)}\sum_{t=1}^{T}(z_{i,t}z_{j,t}-\overline{r}_{ij})^{2},\\
    &= \frac{1}{T(T-1)}(\sum_{t=1}^{T}(z_{i,t}z_{j,t})^{2}-\frac{1}{T}(\sum_{t=1}^{T}z_{i,t}z_{j,t})^{2}).
\end{align}

Finally, the shrinkage parameter is given by

\begin{equation}
    \lambda_{D} = \frac{\Sigma_{i\neq j}\text{Var}(\overline{r}_{ij})}{\Sigma_{i\neq j}\overline{r}^{2}_{ij}}.
\end{equation}
\color{black}

\color{black}
\subsection{Temporal hierarchical reconciliation}
To construct a temporal hierarchy, we first assume a time series has a maximum seasonality of $m$, and a total length $T$ which is a multiple of $m$. For example, we may have monthly observations with a yearly (i.e., $m=12$) seasonality. As above, we express the bottom-level series as $y_{t}$, and we denote the various temporally aggregated series by $x_{j}^{[k]}$, where $\{k_{1},\cdots,k_{p}\}$ are the $p$ factors of $m$ by which we aggregate the time series. That is,

\begin{equation}\label{eq:temporal_agg}
    x_{j}^{[k]}=\sum_{t=(j-1)k+1}^{jk}y_{t},
\end{equation}

where $j=1,\cdots,T/k$. As the observation index $j$ varies with the temporal aggregation level, we define $\tau$ as the index at the highest aggregated level. We define for a given index $\tau$ the vector of values at aggregation level $k$ by

\begin{equation}
    \mathbf{x}^{[k]}_{\tau}=\begin{bmatrix}
        x_{m_{k}(\tau-1)+1}^{[k]}\\
        x_{m_{k}(\tau-1)+2}^{[k]}\\
        \vdots\\
        x_{m_{k}\tau}^{[k]}\\
    \end{bmatrix},
\end{equation}

where $m_{k}=m/k$, $\tau=1,\cdots,N$ and $N=T/m$. We then collect each vector at all aggregation levels for index $\tau$ into

\begin{equation}\label{eq:tempagg}
    \mathbf{x}_{\tau}=\begin{bmatrix}
        x_{\tau}^{[k_{p}]}\\
        \mathbf{x}_{\tau}^{[k_{p-1}]}\\
        \vdots\\
        \mathbf{x}_{\tau}^{k_{1}}
    \end{bmatrix}.
\end{equation}

As with the cross-sectional reconciliation framework, temporal aggregation can be encoded into a summing matrix $\mathbf{S}$ through the relation

\begin{equation}
    \mathbf{x}_{\tau}=\mathbf{S}\mathbf{x}_{\tau}^{[1]},
\end{equation}

where $\mathbf{S}=\begin{bmatrix}
    \mathbf{A}\\
    \mathbf{I}
\end{bmatrix}$, and $\mathbf{A}$ is the aggregation matrix that encodes the factors $k_{1},\cdots,k_{p}$.

However, this hierarchy may not be able to be diagrammatically expressed simply as a single tree as in Figure \ref{fig:hierarchy}. This is because temporal aggregation can be performed at multiple levels which are not additive up the hierarchy. For example, suppose we aggregate at levels $m=2,3,6$. Every two bottom-level observations will aggregate into the $m=2$ observation, but every three bottom-level observations will aggregation into one $m=3$ observations. Similarly every three $m=2$ observations will aggregate to the top-level $m=6$ observation, and every two $m=3$ observations will aggregate to the top-level observation. However, as $2$ and $3$ are relatively prime, this hierarchy cannot be expressed as a single tree. Nonetheless, as long as the highest-level temporal aggregation is a factor of $n$, that is $\frac{m_{0}}{n}\in\mathbb{R}$, the time series can be analyzed through the summation matrix. Uncertainty arises when the temporal aggregation levels are not coprime, as then there does not exist a unique hierarchy. This was first stated in \cite{AthanasopoulosGeorge2017Fwth}, and proven in \cite{YangDazhi2017RsfT}. Nonetheless, once a hierarchy is chosen (if required), the approach of temporal hierarchical reconciliation can become analogous to cross-sectional hierarchical reconciliation. That is, the $\mathbf{G}$ matrix which satisfies Equation \ref{eq:G_matrix} is computed through estimating $\mathbf{W}_{h}$. However, methods of estimating $\mathbf{W}_{h}$ can differ from cross-sectional hierarchies.

\subsection{Cross-temporal reconciliation}

We now consider a scenario in which the hierarchy of consideration has both cross-sectional and temporal aggregations. We denote by $\mathbf{y}_{t}$ the observations at the most temporally disaggregated level, but including all cross-sectional levels of the hierarchy, as defined above. Let $y_{i,t}$ be the $i$th element of this vector, with $i=1,\cdots,n$. We now define the vector of elements \textcolor{black}{$\mathbf{x}_{i,\tau}$} to include all temporally aggregated elements, in accordance with Equation \ref{eq:tempagg}:

\begin{equation}
    \mathbf{x}_{i,\tau}=\begin{bmatrix}
        x_{i,\tau}^{[k_{p}]}\\
        \vdots\\
        \mathbf{x}_{i,\tau}^{[k_{p}]}
    \end{bmatrix},
\end{equation}

where $\tau$ is the temporal index taken from the highest temporally-aggregated level as defined above. Finally, we may combine all cross-sectionally and temporally aggregated observations into a single vector, indexed by $\tau$:

\begin{equation}
    \mathbf{x}_{\tau}=\begin{bmatrix}
        \mathbf{x}_{1,\tau}\\
        \vdots\\
        \mathbf{x}_{n,\tau}
    \end{bmatrix}.
\end{equation}

The relationship between the bottom level cross-sectionally and temporally disaggregated series $\mathbf{b}_{\tau}=\begin{bmatrix}
    \mathbf{x}_{1,\tau}^{[1]} \cdots \mathbf{x}_{n,\tau}^{[1]}
\end{bmatrix}'$ and the total observation vector $\mathbf{x}_{\tau}$ can be neatly described using the tensor product of the cross-sectional and temporal summing matrices:

\begin{equation}\label{eq:crosstemporal_S}
    \mathbf{x}_{\tau}=(\mathbf{S}_{cs}\otimes\mathbf{S}_{te})\mathbf{b}_{\tau}=\mathbf{S}_{ct}\mathbf{b}_{\tau},
\end{equation}

where $\mathbf{S}_{cs},\mathbf{S}_{te}$ and $\mathbf{S}_{ct}$ are the cross-sectional, temporal and cross-temporal summing matrices respectively. Cross-temporal forecast reconciliation can be performed through different methods. One can sequentially reconcile forecasts first across the cross-sectional hierarchy and then across the temporal hierarchy, or vice versa. However, this does not necessarily preserve coherency across the entire cross-temporal hierarchy \cite{kourentzes2019cross}. Heuristic reconciliation refers to a technique in which temporal reconciliation at each cross-sectional level is first performed \cite{kourentzes2019cross}. Projection matrices onto the cross-sectionally coherent subspace are then computed at each temporal aggregation level. Finally, the average projection matrix across all temporal levels is computed and is applied to the temporally reconciled based forecasts. This averaging ensures coherency over the entire cross-temporal hierarchy. Iterative reconciliation can also be performed, in which reconciliation across the cross-sectional and temporal dimensions are performed iteratively until a convergence threshold is met \cite{DiFonzoTommaso2023CfrO}. Finally, one can consider the entire cross-temporal hierarchy simultaneously, that is, $\mathbf{S}_{ct}$, as expressed in Equation \ref{eq:crosstemporal_S}. The optimal reconciliation method then becomes analogous to the cross-sectional case (c.f. Equation \ref{eq:G_matrix}), i.e., finding the matrix $\mathbf{G}$ that minimizes the trace of $\mathbf{V}_{h}$.

One distinguished method in literature which is computationally tractable and has empirically performed well to estimate $\mathbf{W}_{h}$ is optimal combination series variance scaling (oct-wlsv). In oct-wlsv, we set 

\begin{equation}
    \mathbf{W}_{h}=\mathbf{P}'\hat{\boldsymbol{\Omega}}_{wlsv}\mathbf{P},
\end{equation}

where $\mathbf{P}$ is the commutation matrix such that $\mathbf{P}\text{vec}(\mathbf{Y})=\text{vec}(\mathbf{Y}')$ with $\mathbf{Y}$ as a matricization of $\mathbf{x}_{\tau}$ defined in \cite{DiFonzoTommaso2023CfrO}, and $\hat{\boldsymbol{\Omega}}_{wlsv}$ is a diagonal matrix “which contains estimates of the in-sample one-step-ahead error variances across each level” \cite{AthanasopoulosGeorge2017Fwth}.

\section{Base forecasting methods}
\label{sec:base}
\textcolor{black}{\subsection{Linear regression}}
\color{black}
Linear regression (LR) is a statistical model in which output variables are predicted based on a linear combination of the input variables. It remains useful across a variety of statistical modeling scenarios due to its simplicity, interpretability, and ability to provide insights into the relationship between variables, making it a versatile tool for both exploratory analysis and predictive modeling. In LR, given inputs $\mathbf{x}=(1,x_{1},\cdots,x_{p-1})'$, we assume the output can be represented through

\begin{align}
    y&=\mathbf{x}'\boldsymbol{\beta}+e,\\
    y&=\beta_{0}+\beta_{1}x_{1}+\cdots+\beta_{p}x_{p} + e,
\end{align}

where $\beta_{i}\in\mathbb{R}$ are the weighting coefficients of the linear regression, with $i\in[0,p]$ and $e$ is the error which deviates the output from the linear response to its input. Linearity of the model allows for analytic computation of maximum likelihood estimates of these weighting coefficients. The ordinary least squares (OLS) solution can determine the weighting coefficients which minimize the sum of the square of the residuals under the following assumptions:

\begin{enumerate}
    \item The outputs and inputs follow a linear relationship, with the error term having an expected value $\mathbb{E}[e]=0$.
    \item Each observation of inputs and output $(\mathbf{x}_{i},y_{i})$ are drawn from independent and identically distributed (i.i.d.) random variables.
    \item The errors are additive Gaussian noise and uncorrelated with all inputs.
    \item The variance of the errors are constant.
    \item None of the inputs are perfectly collinear.
\end{enumerate}

Under these assumptions, the OLS solution to the weighting coefficients is given by

\begin{equation}
    \hat{\boldsymbol{\beta}}=(\mathbf{X}'\mathbf{X})^{-1}\mathbf{X}'\mathbf{y},
\end{equation}

where $\mathbf{y}=[y_{1},\cdots,y_{n}]'$ are the historical output observations, $\mathbf{X}=[\mathbf{1},\mathbf{x}_{1},\cdots,\mathbf{x}_{p}]$ are the historical input observations, with $\mathbf{1}$ a $(n\times1)$ column vector of $1$'s to allow for the constant term in the linear regression.
\color{black}

\subsection{LightGBM}
\color{black}
Gradient Boosting Decision Trees (GBDT) is a machine learning ensemble technique used for both regression and classification tasks. It builds a predictive model in the form of an ensemble of weak learners (decision trees) and combines them to create a strong predictive model. The fundamental idea behind gradient boosting is to sequentially add weak learners to the ensemble, each correcting the errors of its predecessor. LightGBM, proposed in 2016, builds on GBDT to increase parallel training efficiency, training speed and reduce memory consumption \cite{ke2017lightgbm}. LightGBM empirically performs well in various forecasting scenarios, including making up the majority of the winning algorithms of the M5 competition \cite{MAKRIDAKIS20221346}.

In decision tree regression, the space of all predictor variable values $x$ is partitioned into numerical values given by the leaves of the tree, i.e., terminal nodes. That is, a constant $\gamma_{j}$ is associated with each each region $R_{j}$, with the prediction rule as

\begin{equation}
    x\in R_{j} \implies f(x)=\gamma_{j},
\end{equation}

in which we partition the predictor variable value space into $J$ regions: $j=1,\cdots,J$. Our regression tree can be expressed as

\begin{equation}
    T(x;\Theta)=\sum_{j=1}^{J}\gamma_{j}I(x\in R_{j}),
\end{equation}

where $I(x\in R_{j})$ is the indicator function and $\Theta=\{R_{j},\gamma_{j}\}$ are the parameters associated with the tree. These are set on a decision tree by specifying features such as the maximum depth of the tree, the number of leaves, the minimum data in each leaf, etc. Training the tree then corresponds to minimizing a loss function:

\begin{equation}
    \hat{\Theta}=\operatorname*{argmin}_\Theta \sum_{j=1}^{J}\sum_{\mathbf{x}_{i}\in R_{j}}L(y_{i},\gamma_{j}),
\end{equation}

where $L(y_{i},\gamma_{j})$ is some loss function that quantifies how much we penalize the difference between the actual value $y_{i}$ and the predicted value $\gamma_{j}$. An ensemble of decision trees is given by

\begin{equation}
    f_{M}(x)=\sum_{m=1}^{M}T(x;\Theta_{m}),
\end{equation}

that is, a linear combination of the component decision trees indexed from $m=1,\cdots,M$. Given one decision tree and its predicted values $f(x_{i})$ of the total training data points $x_{i}$, with $ i=1,\cdots,N$, we may consider the vector of predictions

\begin{equation}
    \mathbf{f}=\{f(x_{1}),\cdots,f(x_{N})\}.
\end{equation}

Given this vector, we may compute the components of the gradient of the loss function with respect to these predicted values through

\begin{equation}
    g_{i}=\left[ \frac{\partial L(y_{i},f(x_{i}))}{\partial f(x_{i})} \right].
\end{equation}

In GBDT, we begin with a single decision tree and compute $\mathbf{f}_{0}$. We then iterate the following steps indexed by $m=1,\cdots,M$.

\begin{enumerate}
    \item Compute the gradient and take its negative (i.e. $-\mathbf{g}_{m}$),  to find the direction of greatest decrease in the loss function.
    \item Fit regression trees to the components of this vector to get the new regions $R_{jm}$, with $j=1,\cdots,J_{m}$.
    \item For each region, compute the new constant $\gamma_{jm}$, through\footnote{\textcolor{black}{Note the following regression trees are fit to the residuals of the previous, thereby correcting the errors of previous weak learners.}}
    \begin{equation}
        \gamma_{jm} = \operatorname*{argmin}_\gamma \sum_{\mathbf{x}_{i}\in R_{jm}}L(y_{i},f_{m-1}(x_{i})+\gamma).
    \end{equation}
    \item Update the $m$th decision tree using
    \begin{equation}
        f_{m}(x)=f_{m-1}(x)+\sum_{j=1}^{J_{m}}\gamma_{jm}I(x\in R_{jm}).
    \end{equation}
\end{enumerate}

The final strong learner is given simply by $F(x)=f_{M}(x)$. LightGBM builds on this algorithm through Gradient-based One-Side Sampling (GOSS) and Exclusive Feature Bundling (EFB) \cite{ke2017lightgbm}. GOSS performs downsampling by discarding data instances with small gradients to reduce computational overhead. EFB is a dimensional reduction technique to group features that are mutually exclusive, that is, features that are not simultaneously zero. Using these two techniques, it has been shown empirically that LightGBM can offer training speeds of up to 20 times that of conventional GBDT with comparable accuracy \cite{ke2017lightgbm}.
\color{black}

\section{Data and experimental setup}\label{sec:data}
\subsection{Data}\label{subsec3.1}
\begin{figure}[H]
\centering
\includegraphics[scale=0.8]{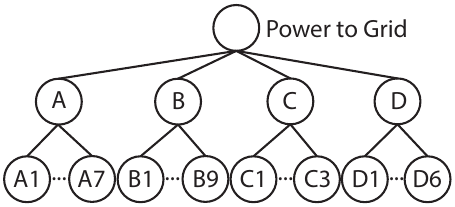}\qquad\qquad\includegraphics[scale=0.8]{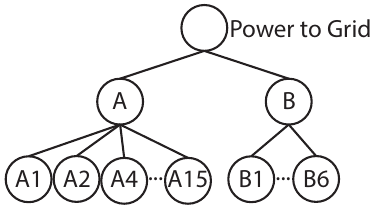}
\caption{Cross-sectional hierarchy of turbines of Data Set A (left) and Data Set B (right)}\label{fig:fig1}
\end{figure}

This work was aimed at investigating the reconciliation of hierarchical time series data relating measured wind speed and \textcolor{black}{energy} generated from wind farms. The objective was to increase the forecasting accuracy across all levels, through comparing different reconciliation approaches. In addition, forecasts were desired at 10-minutely, 20-minutely, 30-minutely, and 1-hourly periods. Two data sets were used, denoted Data Set A and Data Set B. Data Set A comprised 25 wind farm sensors located across a single wind farm in the UK. The farm included four different models of wind turbine which were denoted by A, B, C and D. Data Set B comprised 20 wind farm sensors located across 2 wind farms in Kelmarsh and Penmanshiel in the UK \cite{plumley_charlie_2022_5841834, plumley_charlie_2022_5946808}. Each farm used a different model of wind turbine, which were denoted A and B. The cross-sectional hierarchies of Data Set A and B are shown in Figure \ref{fig:fig1}. \textcolor{black}{In Data Set B, turbine A3 is missing, as the data is unavailable from the original data set \cite{plumley_charlie_2022_5946808}.}

For both data sets, each turbine was equipped with a sensor that measured in 10-minute intervals the average wind speed in m/s, and the average power generated in kW within each interval. Some data were incorrect, typically having a 0 or negative wind speed or power, which was attributed to faults in the turbine, excessively high wind speed, or other issues. \textcolor{black}{To allow additivity in the cross-sectional and temporal hierarchies, we converted the average power to the total energy generated in the 10-minutely period, in units of kWh. Data Set A spanned 1/07/2020 at 00:00 to 30/06/2021 at 23:50, for 365 days of data. Data Set B spanned 01/01/2020 at 00:00:00 to 31-12-2020 at 19:50:00 for 365 days and 20 hours of data.} Both data sets had 3 levels in the hierarchy, with the bottom level representing the individual turbines. The next level up represented each turbine model's aggregated \textcolor{black}{energy} generation. Finally, the root node represented the total \textcolor{black}{energy} generated across all turbines, i.e., the \textcolor{black}{energy} delivered to the electricity grid.

\begin{figure}[H]
\begin{center}
    \includegraphics[width=0.8\linewidth]{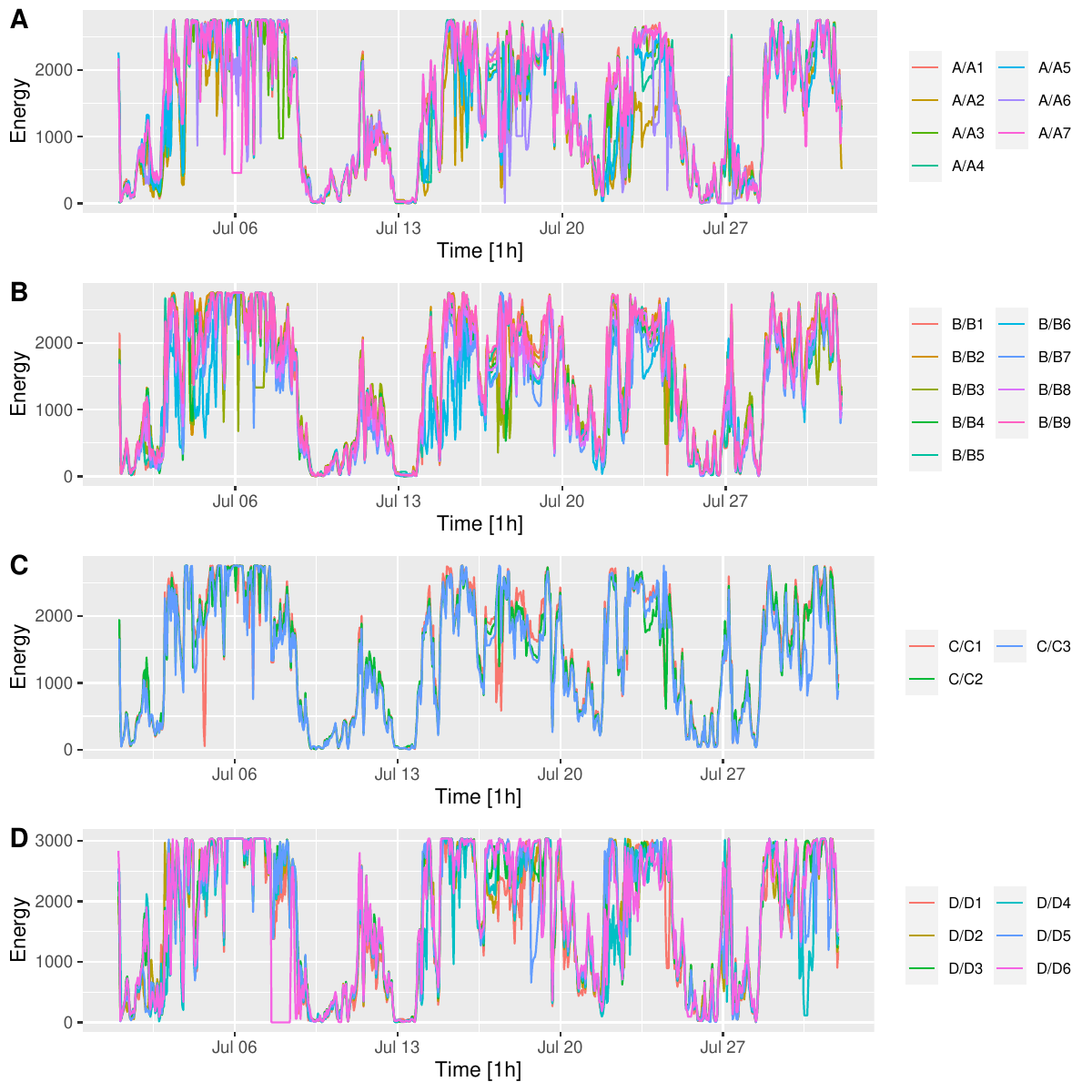}\\
    \end{center}
\caption{\textcolor{black}{Exemplary \textcolor{black}{energy} generation (1-hourly) for each wind turbine over the first month of the training data at Level 2 of the cross-sectional hierarchy, for turbines of model A (A), B (B), C (C) and D (D) of Data Set A}}
\label{fig:example_week}
\end{figure}

\begin{figure}
\begin{center}
    \includegraphics[width=0.8\linewidth]{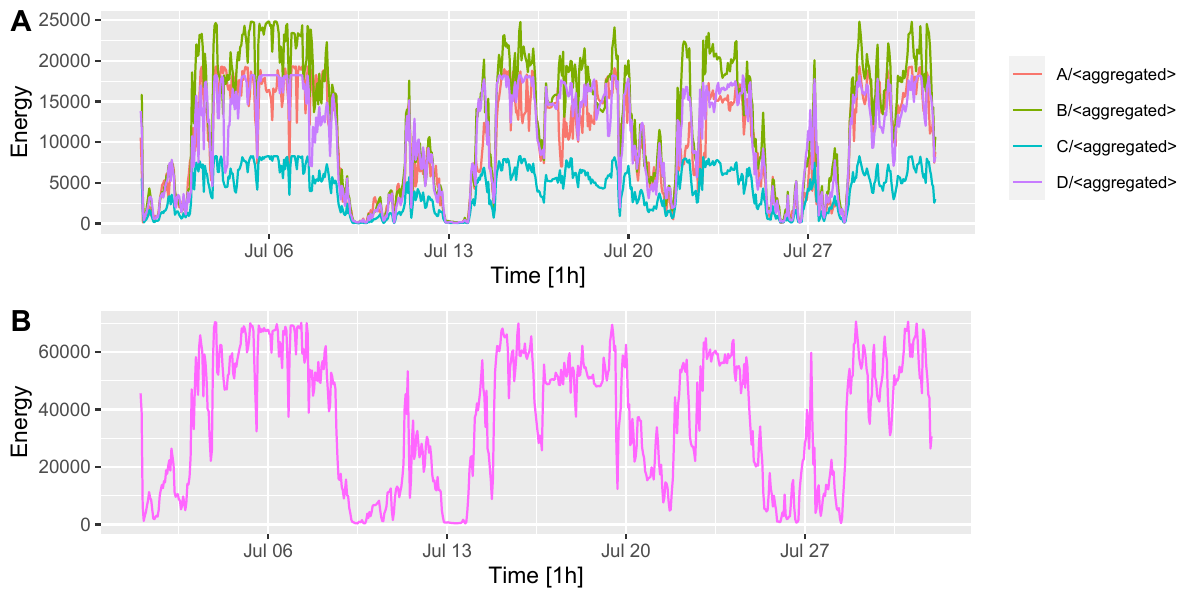}\par
\end{center}
\caption{\textcolor{black}{Exemplary energy generation (1-Hourly) for each wind turbine over the first month of the training data for Level 1 (A) and Level 0 (B) of the cross-sectional hierarchy of Data Set A}}
\label{fig:example_week_agg}
\end{figure}

\begin{figure}
\begin{center}
    \includegraphics[width=0.8\linewidth]{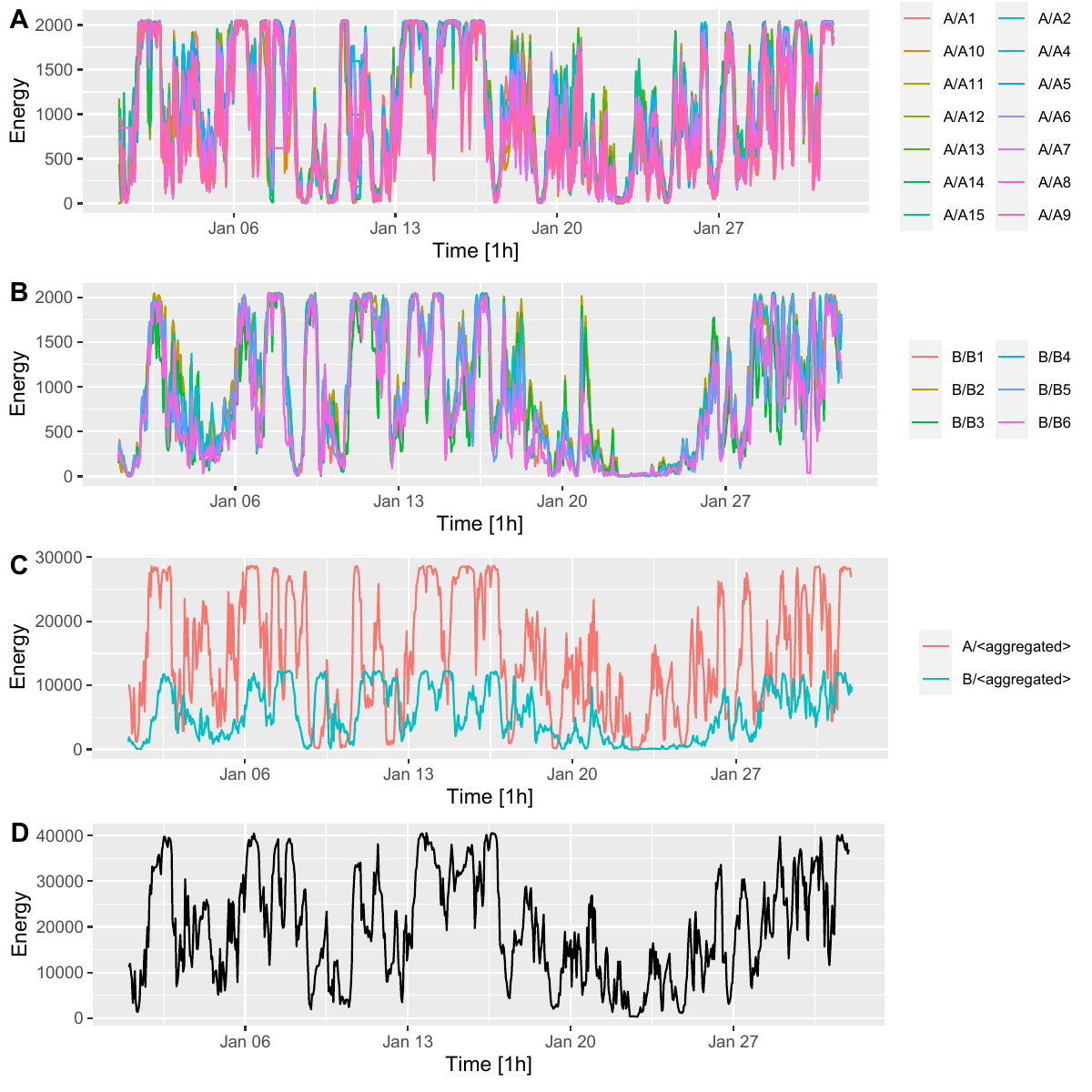}\par
\end{center}
\caption{\textcolor{black}{Exemplary \textcolor{black}{energy} generation (1-Hourly) for each wind turbine over the first month of the training data for turbines of model A (A) and B (B) of Level 2, Level 1 (C) and Level 0 (D) of the cross-sectional hierarchy of Data Set B}}
\label{fig:example_week_dataset_B}
\end{figure}

\textcolor{black}{The data from each cross-sectional node's time series were temporally aggregated in 10-minutely (bottom level), 20-minutely, 30-minutely and 1-hourly (top level) aggregations. Base forecasts were produced at each temporal and cross-sectional aggregation level.} \textcolor{black}{Figures \ref{fig:example_week}-\ref{fig:example_week_dataset_B} present the \textcolor{black}{energy} generated over 1-hourly increments for the bottom level of the hierarchy (Level 2), that is, the individual wind turbines, as well as the total \textcolor{black}{energy} generated for each turbine model (Level 1), and the total \textcolor{black}{energy} delivered to the grid (Level 0) for Data Set A and B.} Because the power \textcolor{black}{(and therefore energy)} generated is approximately linearly proportional to the wind speed, the proximity of the wind turbines from one another leads to similar instantaneous \textcolor{black}{energy} generation, due to similar wind speeds across their locations. In Figures \ref{fig:example_week} and \ref{fig:example_week_dataset_B}, several horizontal lines in time series data can be observed across \textcolor{black}{individual} wind turbines. These lines correspond to periods in which the \textcolor{black}{energy} generated for the specific wind turbine was recorded as negative or NA, and was naive-interpolated from the last non-zero value forwards, until the measurements began again. \textcolor{black}{Recent literature has investigated possible alternatives in wind power forecasting to preprocess unwanted data, including setting them to zero or setting them to the mean value over some fixed period \cite{HanifiShahram2022OWPF}. We have opted for naive-interpolation for computational simplicity while maintaining continuity of the time series.}

\begin{figure}[H]
\begin{multicols}{2}
    \includegraphics[width=\linewidth]{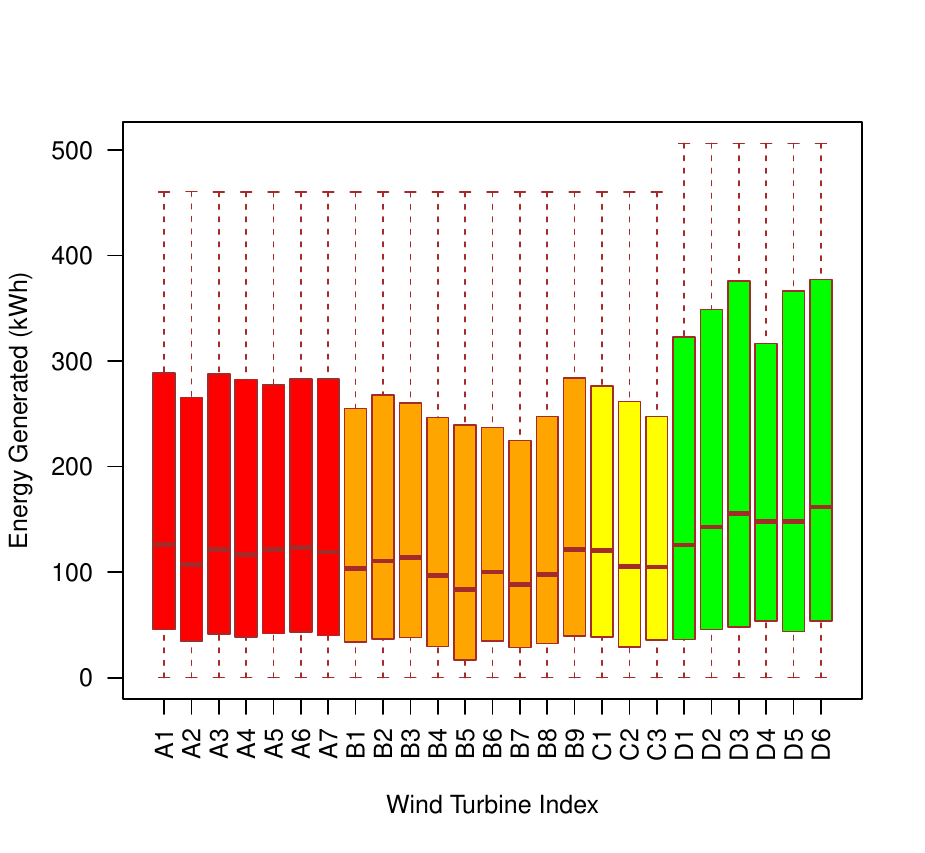}\par
    \includegraphics[width=\linewidth]{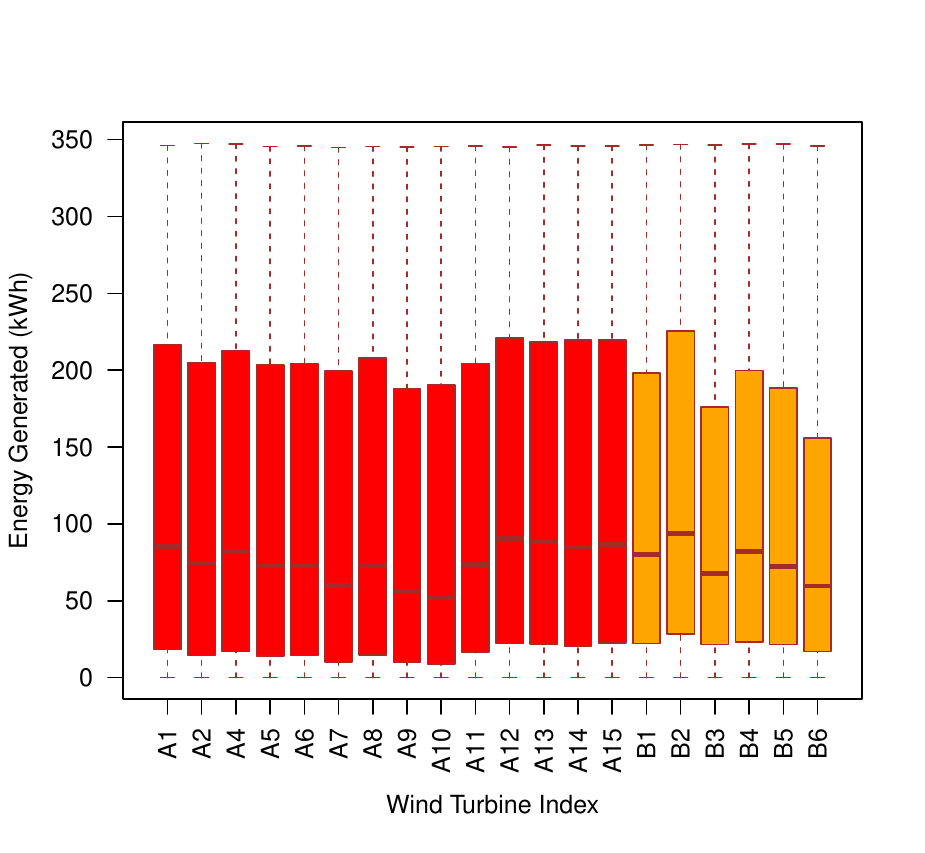}\par
\end{multicols}
\caption{\textcolor{black}{Comparison of 10-minutely energy generation for each wind turbine for Data Set A (left) and B (right)}}
\label{fig:10-min-boxplot}
\end{figure}

A boxplot for the wind turbines in the hierarchy for both data sets is given in Figure \ref{fig:10-min-boxplot}. From Figure \ref{fig:10-min-boxplot}, it can be seen that for Data Set A, turbines of models A-C generated \textcolor{black}{energy} approximately equally, however turbines of model D generated greater \textcolor{black}{energy} than the others on average. Moreover, we observe that the maximum value of the third and first quartiles for turbines of model D was also higher than those of models A-C, indicating different \textcolor{black}{energy} generation capabilities. It can also be seen that for Data Set B, turbine model A contained more turbines than model B. However, all turbines across both models had equal minimum and maximum \textcolor{black}{energies}, and roughly equal \textcolor{black}{energy} generated. This implies that a single model may work well for all turbines, given that they all have similar distributions.

\subsection{Preprocessing}\label{subssec3.2}

Data was preprocessed by using naive interpolation for incorrect values with 0 or negative wind speed or \textcolor{black}{energy}. After, features were extracted to allow for linear and machine learning regression. These features are listed in Table \ref{tab:tab1}. Prior values of wind speed and \textcolor{black}{energy} generation were used as predictors for forecasts, with up to 6 observations prior. A moving average (MA) and moving standard deviation (MSD) of both wind speed and \textcolor{black}{energy} were included to capture the multiple seasonality of the data. The box width for these predictors included $\{2,3,\cdots,6\}$ observations, with equal weights. As described above, there is a strong linear relationship between wind speed and power \textcolor{black}{(energy)} generation. Consequently, the predictors of lagged wind speed and \textcolor{black}{energy} are not linearly independent. However, there is a non-linear relationship for \textcolor{black}{energy} generated by wind-speeds below 2 m/s and above 15 m/s, with the power curve resembling a sigmoid function. Therefore, the variables are not collinear, and hence both wind speed and \textcolor{black}{energy} are included as predictors.

To capture the complex seasonality of the wind speeds, temporal dummy variables were used in the regressions. Seasonal dummy variables included the quarter (i.e., summer, winter, etc.) and the hour of day (in 24-hour time). The month was not included as a seasonal dummy variable due to limitations of the data set. That is, only one year of data was available, so month could not be regressed over if, for example, the training set comprised 11 months, and the testing set comprised 1 month.

\begin{table}[H]\footnotesize
\begin{center}
\caption{Features for time series regression}\label{tab:tab1}%
\centerline{\begin{tabular}{@{}ll@{}}
\toprule
Feature & Number of Predictors\\
\midrule
(1,2,$\cdots$,6)-observation(s) prior wind speed&6\\
(1,2,$\cdots$,6)-observation(s) prior \textcolor{black}{energy} generation&6\\ 
Wind speed moving average (b=2,3,$\cdots$,6)&5\\ 
Wind speed moving standard deviation (b=2,3,$\cdots$,6)&5\\
\textcolor{black}{Energy} moving average (b=2,3,$\cdots$,6)&5\\ 
\textcolor{black}{Energy} moving standard deviation (b=2,3,$\cdots$,6)&5\\
Quarter&3\\ 
Hour of the Day&23\\
\bottomrule
\end{tabular}}
\end{center}
\end{table}
\unskip

\color{black}
The moving average (MA) with box width $b$ at time $k$ of the wind speed is given by

\begin{equation}\label{eq:eq1}
    W_{MA,b}(k) = \frac{1}{b}\sum_{n=k-1-b}^{k-1}W(n),
\end{equation}

where $W(n)$ represents the wind speed at time index $n$. The moving standard deviation (MSD) of box width $b$ at time $k$ of the wind speed is given by

\begin{equation}\label{eq:eq2}
    W_{MSD,b}(k) = \sqrt{\frac{\sum_{n=k-1-b}^{k-1}(W(n)-\frac{1}{b}\sum_{p=k-1-b}^{k-1}W(p))^{2}}{b}}.    
\end{equation}

The moving average of box width $b$ at time $k$ of the energy generated is given by

\begin{equation}\label{eq:eq3}
E_{MA,b}(k) = \frac{1}{b}\sum_{n=k-1-b}^{k-1}E(n),\\
\end{equation}

where $E(n)$ represents the generated energy at time index $n$. The moving standard deviation of box width $b$ at time $k$ of the energy generated is given by

\begin{equation}\label{eq:eq4}
    E_{MSD,b}(k) = \sqrt{\frac{\Sigma_{n=k-1-b}^{k-1}(E(n)-\frac{1}{b}\sum_{p=k-1-b}^{k-1}E(p))^{2}}{b}}
\end{equation}
\color{black}

\subsection{Base forecasts}
The 58 features as shown in Table \ref{tab:tab1} were used in producing base forecasts. \textcolor{black}{Naive forecasts were produced using the Fable package} \textcolor{black}{\cite{fable}}. Linear regression was also performed using the Fable package. Gradient boosting was performed using the LightGBM package \textcolor{black}{\cite{lightgbm}}. A custom wrapper was written for LightGBM to act as a model to allow hierarchical reconciliation through the fabletools package in the R programming language \cite{fabletools}. The three base forecasting methods were then classified as naive, linear regression and gradient boosting.
Base forecasts used linear regression (LR) and light gradient boosting machine learning (GB), with features given in Table \ref{tab:tab1}. One hour ahead was forecast at each step, with the 10-minutely data having 6-steps-ahead, the 20-minutely data having 3-steps-ahead, the 30-minutely data having 2-steps-ahead and the 1-hourly data having 1-step-ahead. These one-hour-ahead forecasts were then cross-temporally reconciled using the algorithms described above, to compare the accuracy of the different cross-temporal reconciliation algorithms.

 The hyperparameters of the GB model were first optimized using a grid search algorithm, \textcolor{black}{with a data set split of 80\% training data, 10\% test data and 10\% validation data. Hyperparameters optimized included max number of leaves in one tree (\textit{num\_leaves}), the max depth for a tree model (\textit{max\_depth}), the learning rate (\textit{learning\_rate}) and the minimal number of data in one leaf (\textit{min\_data\_in\_leaf})}. The optimal hyperparameters for each temporal aggregation is given in Table \ref{tab:hyperparams}.

\begin{table}[H]\footnotesize
\color{black}
\begin{center}
\caption{Optimal hyperparameters for all temporal aggregations for GB model\label{tab:hyperparams}}
		\centerline{\begin{tabular}{@{}llllll@{}}
			\toprule
			Data Set & Hyperparameters & 10-minutely & 20-minutely & 30-minutely & 1-hourly\\
   			\midrule
A & num\_leaves & 100 & 120 & 80 & 100\\
 &max\_depth & 9 & 9 & 8 & 8\\
 &learning\_rate & 0.2 & 0.2 & 0.1 & 0.2\\
 &min\_data\_in\_leaf & 250 & 325 & 200 & 200\\
\midrule
B & num\_leaves         & 130  & 90  & 50  & 80  \\
& max\_depth            & 8   & 9   & 10   & 8   \\
& learning\_rate      & 0.1 & 0.125 & 0.15 & 0.175 \\
& min\_data\_in\_leaf & 200 & 375 & 200 & 175 \\
			\bottomrule
		\end{tabular}}
\end{center}
\color{black}
\end{table}

\subsection{Forecast reconciliation}
After generating base forecasts with LR and GB, we use several methods for \textcolor{black}{cross-sectional reconciliation of} time series. These include TD, BU, MO, and COM methods with OLS and MinT estimators. \textcolor{black}{We use the \textit{hts} package in R for this purpose \cite{hts}}. \textcolor{black}{For cross-temporal reconciliation, we used six different algorithms, including (1) cross-temporal bottom-up reconciliation (ct(bu)), (2) forecast reconciliation through temporal hierarchies for all time series using series autocovariance (t-acov) \cite{NystrupPeter2020Thwa}, (3) heuristic first-temporal-then-cross-sectional cross-temporal reconciliation using temporal series variance and cross-sectional shrinkage (tcs-wlsv-shr) \cite{kourentzes2019cross}, (4) heuristic first-cross-sectional-then-temporal cross-temporal reconciliation using temporal autocovariance and cross-sectional shrinkage (cst-shr-acov) \cite{DiFonzoTommaso2023CfrO}, (5) iterative cross-temporal reconciliation using temporal series autocovariance and cross-sectional shrinkage (ite-acov-shr) \cite{DiFonzoTommaso2023CfrO} and (6) optimal cross-temporal reconciliation using series variance scaling (oct-wlsv) \cite{DiFonzoTommaso2023CfrO}. \textcolor{black}{The t-acov reconciliation method only considers temporal reconciliation, so will not ensure cross-temporal coherency. We have included this method as literature has suggested more computationally complex cross-temporal reconciliation may not offer significant accuracy gains compared with temporal-only reconciliation \cite{ATHANASOPOULOS2023}.} Cross-temporal reconciliation was performed using the \textit{FoReco} package \textcolor{black}{\cite{foreco}}.}

\textcolor{black}{Forecast reconciliation requires an estimate of the h-steps-ahead error covariance matrix, which is computed using the in-sample forecast residuals. The desired forecasts in this work are 1-hour-ahead, with temporal granularities of six 10-minutely, three 20-minutely, two 30-minutely and one 1-hourly forecasts. For the 10-minutely forecasts, we train 6 different LR/GB models that each produce a single forecast. The 2-,3-,…,6-steps-ahead models are trained on lagged predictors, such that forecasts are produced using only data that would be available in a practical forecasting scenario. Similarly, for the 20- and 30-minutely forecasts, we have 3 and 2 separate models trained, respectively. 
In cross-temporal reconciliation, typically a single model is fit at each temporal granularity (e.g., ARIMA \cite{DiFonzoTommaso2023CfrO, GIROLIMETTO2023} or ETS \cite{DIFONZO202313}), and the in-sample residuals from each model are used to estimate the error covariance matrix for reconciliation. We may consider our multi-model forecasts as originating from a single \textit{effective} model also.}

\textcolor{black}{For example, suppose each of the six 10-minutely models are indexed by $i=1,…,6$, with forecasts at time $t$ given by $\mathbf{y}^{i}_{t}$. Then our effective model is defined through forecasts given by}

\textcolor{black}{
\begin{equation}
    \mathbf{y}_{t} =
\left\{
	\begin{array}{ll}
		\mathbf{y}^{1}_{t} & \mbox{if } (t-1)\text{mod}6=0, \\
            \mathbf{y}^{2}_{t} & \mbox{if } (t-2)\text{mod}6=0, \\
            & \vdots \\
            \mathbf{y}^{6}_{t} & \mbox{if } (t-6)\text{mod}6=0.
	\end{array}
\right.
\end{equation}}

\textcolor{black}{That is, the first forecast is taken from model 1, the second forecast is taken from model 2, etc. When we forecast 1-hour-ahead with this effective model, we produce 1 forecast from each of the individual models as expected. It follows that the residuals from this effective model are given by}

\textcolor{black}{
\begin{equation}
    \hat{e}_{t}(h) = 
\left\{
	\begin{array}{ll}
		\hat{e}^{1}_{t}(h) & \mbox{if } (t-1)\text{mod}6=0, \\
            \hat{e}^{2}_{t}(h) & \mbox{if } (t-2)\text{mod}6=0, \\
            & \vdots \\
            \hat{e}^{6}_{t}(h) & \mbox{if } (t-6)\text{mod}6=0.
	\end{array}
\right.
\end{equation}}

\textcolor{black}{The same equations follow for the 20- and 30-minutely models also, using mod3 and mod2, respectively. Finally, to reconcile, we interleave the residuals from the individual models in the same order that the forecasts will be produced. Ensuring the same order will also allow temporal correlations within the residuals to be captured during reconciliation.}

\subsection{Forecast evaluation}
Accuracy was evaluated on a rolling origin basis, whereby the origin was rolled one hour forwards each step, and an hour of forecasts were produced at each temporal granularity. As the time-series data had an additive hierarchical structure, the magnitude of forecasts at each level differed. To account for these differing magnitudes, average relative accuracy indices were used in evaluating forecast accuracies \cite{DiFonzoTommaso2023CfrO}. Average relative accuracy indices were used in favour of scaled errors, such as MASE or RMSSE \cite{HYNDMAN2006679}, to minimize the bias towards overrating the performance of a benchmark forecast as a result of arithmetic averaging, and to ensure accuracy indices were robust to outliers, in the case of dividing by small benchmark error measures. The average relative root mean squared error (AvgRelRMSE) was used for comparison, in which high weight is given to larger errors due to squaring.

\textcolor{black}{Suppose for each time $t$,} we first represent the forecast error  through

\begin{equation}
    \hat{e_{j}} = y_{j}-\hat{y}_{j}.
\end{equation}

Then, for a \textcolor{black}{reconciled forecast}, the RMSE is computed through

\begin{equation}
    RMSE = \sqrt{\frac{1}{n}\mathlarger{\sum}_{j=1}^{n}(\hat{e_{j}})^{2}}.
\end{equation}

For each time series, we can then scale it by a benchmark forecast error, for example, an $h$-steps-ahead naive forecast. With a benchmark error of $\hat{e_{b,j}}$, we can compute the relative RMSE through

\begin{equation}
    RelRMSE = \frac{\sqrt{\frac{1}{n}\mathlarger{\sum}_{j=1}^{n}(\hat{e_{j}})^{2}}}{\sqrt{\frac{1}{n}\mathlarger{\sum}_{j=1}^{n}(\hat{e_{b,j}})^{2}}}.
\end{equation}

Finally, to compute the average RelRMSE of a set of forecasts in a hierarchy, either cross-sectional, temporal, or cross-temporal, we simply take the geometric mean of the set of relative errors. That is, for a set of \textcolor{black}{reconciled} time series indexed by $ts=\{1,2,\cdots,q\}$, the average RelRMSE is computed through

\begin{equation}
    AvgRelRMSE = (\mathlarger{\prod}_{ts=1}^{q} \frac{\sqrt{\frac{1}{n}\mathlarger{\sum}_{j=1}^{n}(\hat{e^{ts}_{j}})^{2}}}{\sqrt{\frac{1}{n}\mathlarger{\sum}_{j=1}^{n}(\hat{e^{ts}_{b,j}})^{2}}})^{\frac{1}{q}}.
\end{equation}

For further details on the experiments, please see the R code publicly available at \url{https://github.com/englishlukeh/WindPowerForecasting}.

\section{Empirical results and discussion}\label{sec:results}

Using a training set of 90\%, and a testing set of 10\% of the total 1-year data set, time series cross-validation (TSCV) was used to evaluate the effectiveness of the forecasting. We present the results for cross-sectional and cross-temporal in sections \ref{cross-sec} and \ref{cross-temp}, respectively.

\subsection{Cross-sectional reconciliation}\label{cross-sec}
The results of the AvgRelRMSE, after cross-sectional reconciliation, are given in Tables \ref{tab:results1} and \ref{tab:results2}. The AvgRelRMSE at levels 2 (L2), 1 (L1), 0 (L0), and across all levels of the hierarchy, that is, the total geometric mean (GM), are provided. Values greater than one indicate poorer performance measured through the average relative error index than the naive $h$-steps-ahead benchmark, \textcolor{black}{and are indicated in red}. Values less than one indicate \textcolor{black}{better} performance. In these tables, LR denotes base forecasts produced through linear regression, and GB denotes base forecasts produced through LightGBM. The cross-sectional reconciliation method used is denoted through BU (bottom-up), TD (top-down), MO (middle-out) OLS (ordinary least squares) and MinT-Shr (minimum trace with shrinkage estimator).

At the 10-minutely temporally-disaggregated level, for Data Set A, \textcolor{black}{LR-MinT-Shr outperformed other methods} for wind turbine forecasting (i.e., Level 2) and across all levels of the hierarchy. However, GB-MinT-Shr performed better for aggregated turbine forecasting (Level 1), and GB-OLS performed better for forecasting \textcolor{black}{energy} delivered to the grid (Level 0). \textcolor{black}{While GB performed well on Data Set A, GB was unable to outperform the naive model on Data Set B across all time horizons. The discrepancy between GB performance across Data Set A and B indicates that the inherent nature of data has been pivotal in obtaining such accuracies.} For Data Set B, LR-TD performed best across all levels. LR base forecasts were approximately $1.8\%$ less accurate for Data Set B compared to Data Set A, and GB base forecasts were approximately $8\%$ less accurate. The best reconciliation method was $1.9\%$ more accurate for Data Set A than Data Set B. This suggests greater forecastability for Data Set A. We also observe that the GB model was significantly less accurate for Data Set B \textcolor{black}{and performed worse than a naive model}, indicating that simple methods like LR can be more accurate than more complex methods such as GB when data are more forecastable.

At the 20-minutely level, for Data Set A, \textcolor{black}{LR-MinT-Shr outperformed other methods} for wind turbine forecasting, aggregated turbine model forecasting and across all levels of the hierarchy. GB-OLS performed better for forecasting \textcolor{black}{energy} delivered to the grid. For Data Set B, LR-TD outperformed other methods across all levels. 

At the 30-minutely level, for both Data Set A and B, \textcolor{black}{LR-MinT-Shr} performed best for wind turbine forecasting, aggregated turbine model forecasting and across all levels of the hierarchy. The base forecast without reconciliation (i.e., LR) at the root node, that is, \textcolor{black}{energy} delivered to the grid, performed best.

At the 1-hourly level, for Data Set A, LR-MinT-Shr performed best for wind turbine forecasting, aggregated turbine model forecasting and across all levels of the hierarchy. The base forecast without reconciliation (i.e., LR) at the root node was the top-performing method. For Data Set B, \textcolor{black}{LR-MinT-Shr outperformed other methods across all levels. As the temporal granularity decreases, the MinT algorithm becomes increasingly better than the other methods across both data sets}.

\begin{table}[H]\footnotesize
\color{black}
\begin{center}
\caption{Forecasting performance of \textcolor{black}{cross-sectional reconciliation} models in the present study at individual temporal aggregations}\label{tab:results1}%
\centerline{\begin{tabular}{@{}ll|llll|llll@{}}
\toprule
Data Set & HF Method         & L2 & L1 & L0 & GM & L2 & L1 & L0 & GM\\
\midrule
        &        &  \multicolumn{4}{c|}{\textbf{10-Minutely}} & \multicolumn{4}{c}{\textbf{20-Minutely}}\\
        \midrule
A & LR          & 0.957       & 0.962 & 0.964 & 0.958 & 0.973    & 0.977 & 0.977 & 0.973 \\
  & LR-BU       & 0.957       & 0.966 & 0.986 & 0.959 & 0.972    & 0.980 & \textcolor{red}{1.001} & 0.974 \\
  & LR-TD       & 0.948       & 0.954 & 0.964 & 0.949 & 0.962    & 0.967 & 0.977 & 0.963 \\
  & LR-MO       & 0.954       & 0.962 & 0.975 & 0.956 & 0.968    & 0.975 & 0.989 & 0.969 \\
  & LR-OLS      & 0.948       & 0.954 & 0.965 & 0.949 & 0.962    & 0.968 & 0.979 & 0.963 \\
  & LR-MinT-Shr & \textbf{0.945}       & 0.950 & 0.964 & \textbf{0.946} & \textbf{0.957}    & \textbf{0.962} & 0.977 & \textbf{0.959} \\
  & GB          & 0.961       & 0.958 & 0.965 & 0.961 & 0.975    & 0.976 & 0.981 & 0.975 \\
  & GB-BU       & 0.960       & 0.956 & 0.970 & 0.960 & 0.974    & 0.972 & 0.987 & 0.975 \\
  & GB-TD       & 0.968       & 0.965 & 0.965 & 0.968 & 0.981    & 0.978 & 0.981 & 0.980 \\
  & GB-MO       & 0.963       & 0.958 & 0.959 & 0.962 & 0.979    & 0.976 & 0.978 & 0.978 \\
  & GB-OLS      & 0.962       & 0.955 & \textbf{0.958} & 0.961 & 0.976    & 0.972 & \textbf{0.975} & 0.975 \\
  & GB-MinT-Shr & 0.956       & \textbf{0.949} & 0.960 & 0.955 & 0.974    & 0.970 & 0.985 & 0.974 \\
  \midrule
B & LR          & 0.975       & 0.981 & \textbf{0.987} & 0.976 & 0.977    & 0.976 & \textbf{0.979} & 0.977 \\
  & LR-BU       & 0.975       & 0.994 & \textcolor{red}{1.010} & 0.978 & 0.977    & 0.987 & \textcolor{red}{1.002} & 0.979 \\
  & LR-TD       & \textbf{0.963}       & \textbf{0.977} & \textbf{0.987} & \textbf{0.965} & \textbf{0.965}    & \textbf{0.971} & \textbf{0.979} & \textbf{0.966} \\
  & LR-MO       & 0.965       & 0.981 & 0.991 & 0.968 & 0.968    & 0.976 & 0.985 & 0.969 \\
  & LR-OLS      & 0.963       & 0.978 & 0.989 & 0.966 & 0.965    & 0.972 & 0.981 & 0.967 \\
  & LR-MinT-Shr & 0.964       & 0.978 & 0.989 & 0.966 & 0.965    & 0.972 & 0.981 & 0.967 \\
  & GB          & \textcolor{red}{1.042}       & \textcolor{red}{1.031} & \textcolor{red}{1.037} & \textcolor{red}{1.041} & \textcolor{red}{1.035}    & \textcolor{red}{1.020} & \textcolor{red}{1.029} & \textcolor{red}{1.033} \\
  & GB-BU       & \textcolor{red}{1.042}       & \textcolor{red}{1.044} & \textcolor{red}{1.107} & \textcolor{red}{1.045} & \textcolor{red}{1.035}    & \textcolor{red}{1.027} & \textcolor{red}{1.075} & \textcolor{red}{1.036} \\
  & GB-TD       & \textcolor{red}{1.010}       & \textcolor{red}{1.019} & \textcolor{red}{1.037} & \textcolor{red}{1.012} & \textcolor{red}{1.011}    & \textcolor{red}{1.008} & \textcolor{red}{1.029} & \textcolor{red}{1.012} \\
  & GB-MO       & \textcolor{red}{1.028}       & \textcolor{red}{1.031} & \textcolor{red}{1.075} & \textcolor{red}{1.030} & \textcolor{red}{1.027}    & \textcolor{red}{1.020} & \textcolor{red}{1.057} & \textcolor{red}{1.028} \\
  & GB-OLS      & \textcolor{red}{1.022}       & \textcolor{red}{1.029} & \textcolor{red}{1.042} & \textcolor{red}{1.023} & \textcolor{red}{1.021}    & \textcolor{red}{1.018} & \textcolor{red}{1.031} & \textcolor{red}{1.021} \\
  & GB-MinT-Shr & \textcolor{red}{1.016}       & \textcolor{red}{1.017} & \textcolor{red}{1.051} & \textcolor{red}{1.017} & \textcolor{red}{1.020}    & \textcolor{red}{1.012} & \textcolor{red}{1.045} & \textcolor{red}{1.020} \\
\midrule
        &        &  \multicolumn{4}{c|}{\textbf{30-Minutely}} & \multicolumn{4}{c}{\textbf{1-Hourly}}        \\
        \midrule
A & LR          & 0.973       & 0.974 & \textbf{0.962} & 0.973 & 0.963    & 0.960 & \textbf{0.941} & 0.962 \\
  & LR-BU       & 0.972       & 0.978 & 0.990 & 0.973 & 0.963    & 0.963 & 0.961 & 0.963 \\
  & LR-TD       & 0.959       & 0.962 & \textbf{0.962} & 0.959 & 0.953    & 0.951 & \textbf{0.941} & 0.952 \\
  & LR-MO       & 0.968       & 0.973 & 0.978 & 0.969 & 0.960    & 0.960 & 0.953 & 0.960 \\
  & LR-OLS      & 0.960       & 0.963 & 0.965 & 0.961 & 0.954    & 0.952 & 0.943 & 0.953 \\
  & LR-MinT-Shr & \textbf{0.954}       & \textbf{0.956} & 0.963 & \textbf{0.955} & \textbf{0.949}    & \textbf{0.946} & 0.942 & \textbf{0.948} \\
  & GB          & 0.977       & 0.976 & 0.979 & 0.977 & 0.976    & 0.968 & 0.969 & 0.975 \\
  & GB-BU       & 0.977       & 0.970 & 0.981 & 0.976 & 0.975    & 0.963 & 0.958 & 0.973 \\
  & GB-TD       & 0.987       & 0.980 & 0.979 & 0.986 & 0.992    & 0.982 & 0.969 & 0.990 \\
  & GB-MO       & 0.982       & 0.976 & 0.974 & 0.981 & 0.980    & 0.968 & 0.953 & 0.978 \\
  & GB-OLS      & 0.982       & 0.973 & 0.973 & 0.980 & 0.985    & 0.971 & 0.962 & 0.982 \\
  & GB-MinT-Shr & 0.972       & 0.965 & 0.972 & 0.971 & 0.989    & 0.980 & 0.976 & 0.987 \\
  \midrule
B & LR          & 0.980       & 0.980 & \textbf{0.979} & 0.980 & 0.984    & 0.983 & 0.981 & 0.984 \\
  & LR-BU       & 0.980       & 0.985 & 0.996 & 0.981 & 0.984    & 0.986 & 0.988 & 0.984 \\
  & LR-TD       & 0.972       & 0.975 & \textbf{0.979} & 0.972 & 0.980    & 0.979 & 0.981 & 0.980 \\
  & LR-MO       & 0.975       & 0.980 & 0.984 & 0.976 & 0.982    & 0.983 & 0.983 & 0.982 \\
  & LR-OLS      & 0.972       & 0.975 & 0.981 & 0.973 & 0.979    & 0.979 & 0.981 & 0.979 \\
  & LR-MinT-Shr & \textbf{0.971}       & \textbf{0.974} & 0.980 & \textbf{0.972} & \textbf{0.977}    & \textbf{0.977} & \textbf{0.979} & \textbf{0.977} \\
  & GB          & \textcolor{red}{1.032}       & \textcolor{red}{1.020} & \textcolor{red}{1.016} & \textcolor{red}{1.030} & \textcolor{red}{1.042}    & \textcolor{red}{1.016} & \textcolor{red}{1.005} & \textcolor{red}{1.038} \\
  & GB-BU       & \textcolor{red}{1.032}       & \textcolor{red}{1.017} & \textcolor{red}{1.057} & \textcolor{red}{1.031} & \textcolor{red}{1.042}    & \textcolor{red}{1.023} & \textcolor{red}{1.053} & \textcolor{red}{1.040} \\
  & GB-TD       & \textcolor{red}{1.015}       & \textcolor{red}{1.011} & \textcolor{red}{1.016} & \textcolor{red}{1.015} & \textcolor{red}{1.017}    & \textcolor{red}{1.008} & \textcolor{red}{1.005} & \textcolor{red}{1.015} \\
  & GB-MO       & \textcolor{red}{1.032}       & \textcolor{red}{1.020} & \textcolor{red}{1.050} & \textcolor{red}{1.032} & \textcolor{red}{1.031}    & \textcolor{red}{1.016} & \textcolor{red}{1.031} & \textcolor{red}{1.030} \\
  & GB-OLS      & \textcolor{red}{1.026}       & \textcolor{red}{1.020} & \textcolor{red}{1.019} & \textcolor{red}{1.025} & \textcolor{red}{1.025}    & \textcolor{red}{1.014} & \textcolor{red}{1.008} & \textcolor{red}{1.024} \\
  & GB-MinT-Shr & \textcolor{red}{1.025}       & \textcolor{red}{1.013} & \textcolor{red}{1.040} & \textcolor{red}{1.025} & \textcolor{red}{1.046}    & \textcolor{red}{1.029} & \textcolor{red}{1.060} & \textcolor{red}{1.045} \\
\bottomrule
\end{tabular}}
\footnotesize{Best solutions in each category are indicated in bold}
\end{center}
\end{table}

For the geometric means across all temporal aggregations, for both Data Set A and B, \textcolor{black}{LR-MinT-Shr} performed best for forecasts of wind turbines, aggregated turbine models and across all levels of the hierarchy. Base forecasts at the root node without reconciliation outperformed other methods. \textcolor{black}{For Data Set B, LR-MinT-Shr demonstrated similar levels of accuracy.} The best-performing reconciliation method increased the accuracy over base forecasts by $1.4\%$ for Data Set A, and $0.9\%$ for Data Set B.

We observe by aggregating data, the performance of TD models is increasing at the top level. However, the improvement is not equal across different levels and data. The accuracy achievement at Level 0 is relatively equal for 10-minutely, 20-minutely, 30-minutely, and hourly data. However, at Level 2 we observe a higher gain for 10-minutely data. This shows that the frequency of data may impact the performance of reconciliation methods. While \textcolor{black}{MinT-Shr} remains a top-performing model across all data and levels, other methods behave differently with TD being better than others. The choice of the forecasting method is also important. We see a different performance for LR and GB models, with LR consistently doing better than GB. This may be different in other data \textcolor{black}{with different frequencies, but in our experiment for high-frequency data starting from 10-minute granularity up to one hour, a simple LR model with a sufficient number of features is able to forecast more accurately than a GB model. Noting that the GB was a winning model in the M5 forecasting competition on low-frequency data (daily granularity), this requires further investigation in future research.}

\begin{table}[H]\footnotesize
\color{black}
\begin{center}
\caption{Forecasting performance of \textcolor{black}{cross-sectional reconciliation} models in the present study across all temporal aggregations}\label{tab:results2}
\centerline{\begin{tabular}{@{}l|llll|llll@{}}
\toprule
         & \multicolumn{8}{c}{\textbf{All temporal aggregations}}                \\
         \midrule
         & \multicolumn{4}{c|}{Data Set A} & \multicolumn{4}{c}{Data Set B} \\
         \midrule
HF Method                 & L2 & L1 & L0 & GM & L2 & L1 & L0 & GM\\
\midrule
LR          & 0.966 & 0.968 & \textbf{0.961} & 0.966 & 0.979 & 0.980 & \textbf{0.982} & 0.979 \\
LR-BU       & 0.966 & 0.972 & 0.984 & 0.967 & 0.979 & 0.988 & 0.999 & 0.981 \\
LR-TD       & 0.955 & 0.959 & \textbf{0.961} & 0.956 & 0.970 & 0.976 & \textbf{0.982} & 0.971 \\
LR-MO       & 0.962 & 0.968 & 0.974 & 0.963 & 0.972 & 0.980 & 0.986 & 0.974 \\
LR-OLS      & 0.956 & 0.959 & 0.963 & 0.957 & 0.970 & 0.976 & 0.983 & 0.971 \\
LR-MinT-Shr & \textbf{0.951} & \textbf{0.953} & 0.961 & \textbf{0.952} & \textbf{0.969} & \textbf{0.975} & 0.982 & \textbf{0.970} \\
GB          & 0.972 & 0.969 & 0.973 & 0.972 & \textcolor{red}{1.037} & \textcolor{red}{1.022} & \textcolor{red}{1.022} & \textcolor{red}{1.035} \\
GB-BU       & 0.972 & 0.965 & 0.974 & 0.971 & \textcolor{red}{1.037} & \textcolor{red}{1.028} & \textcolor{red}{1.073} & \textcolor{red}{1.038} \\
GB-TD       & 0.982 & 0.976 & 0.973 & 0.981 & \textcolor{red}{1.013} & \textcolor{red}{1.011} & \textcolor{red}{1.022} & \textcolor{red}{1.013} \\
GB-MO       & 0.976 & 0.969 & 0.966 & 0.975 & \textcolor{red}{1.030} & \textcolor{red}{1.022} & \textcolor{red}{1.053} & \textcolor{red}{1.030} \\
GB-OLS      & 0.976 & 0.968 & 0.967 & 0.975 & \textcolor{red}{1.023} & \textcolor{red}{1.021} & \textcolor{red}{1.025} & \textcolor{red}{1.023} \\
GB-MinT-Shr & 0.973 & 0.966 & 0.974 & 0.972 & \textcolor{red}{1.027} & \textcolor{red}{1.018} & \textcolor{red}{1.049} & \textcolor{red}{1.027} \\
\bottomrule
\end{tabular}}
\footnotesize{Best solutions in each category are indicated in bold}
\end{center}
\end{table}

\begin{figure}[H]
\centering
\includegraphics[scale=0.6]{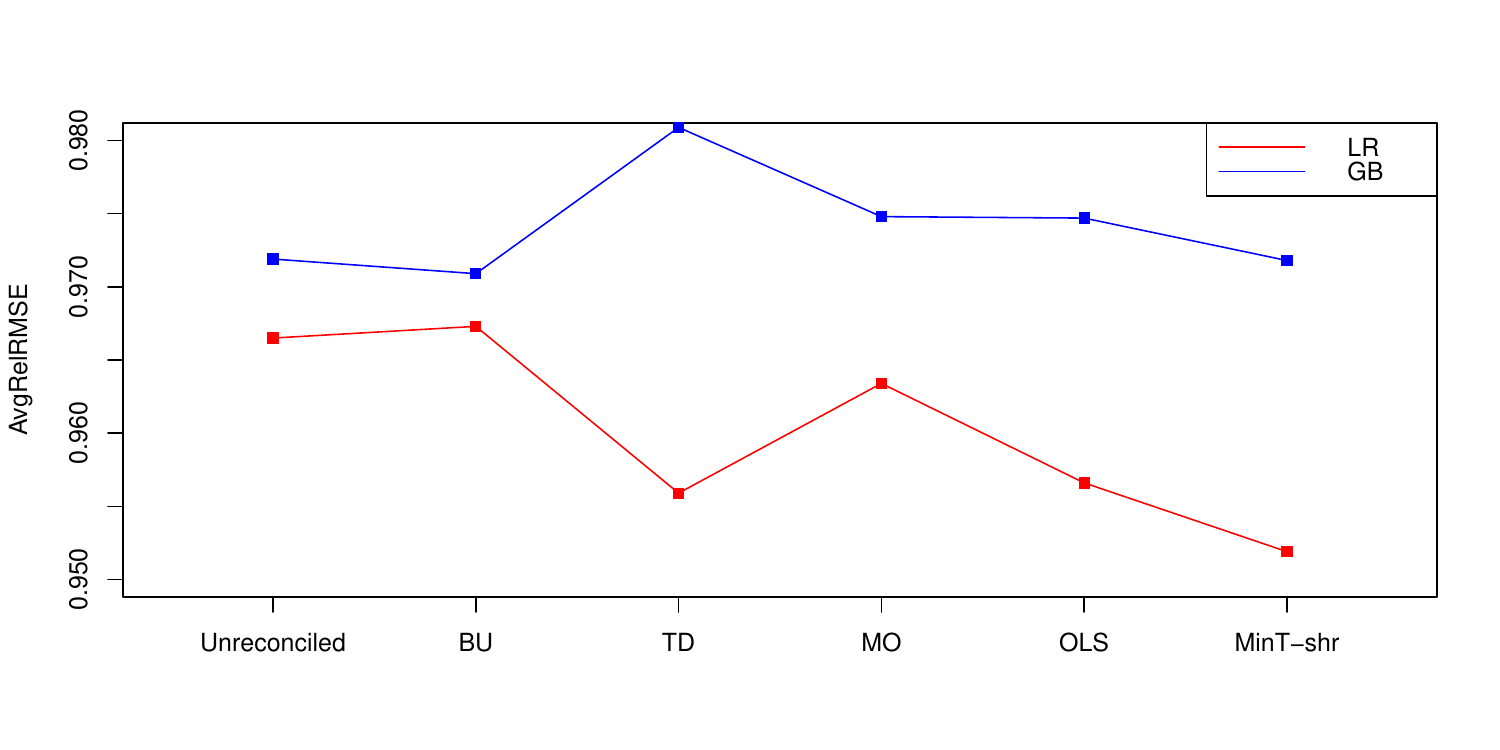}\\
\caption{\textcolor{black}{Forecasting performance of \textcolor{black}{cross-sectional reconciliation} models in the present study across all cross-sectional and temporal aggregations for Data Set A}}
\label{fig:avgrelrmse}
\end{figure}

One dominant trend across all cross-sectional and temporal aggregation levels is the performance of the \textcolor{black}{MinT-Shr} and TD reconciliation algorithms. Additionally, unreconciled base forecasts proved best for the cross-sectional root node, that is, \textcolor{black}{energy} delivered to the grid. Across most levels, LR outperformed GB, which may suggest a local minimum was found during hyperparameter optimization, instead of the global minimum. Nonetheless, GB performed best at the 10- and 20-minutely temporal aggregation levels for \textcolor{black}{energy} delivered to the grid. A comparison of the forecasting performance across all cross-sectional and temporal aggregations is shown in Figure \ref{fig:avgrelrmse}. For both data sets, TD and MinT exhibited the greatest accuracy with LR base forecasts. However, TD performed poorly with GB forecasts for Data Set A, but very well relative to other GB reconciliation methods for Data Set B. This may suggest greater variability in the reconciliation of GB base forecasts when compared with LR base forecasts. However, MinT also performed well with GB for both data sets.

\begin{figure}[H]
\centering
\includegraphics[scale=0.6]{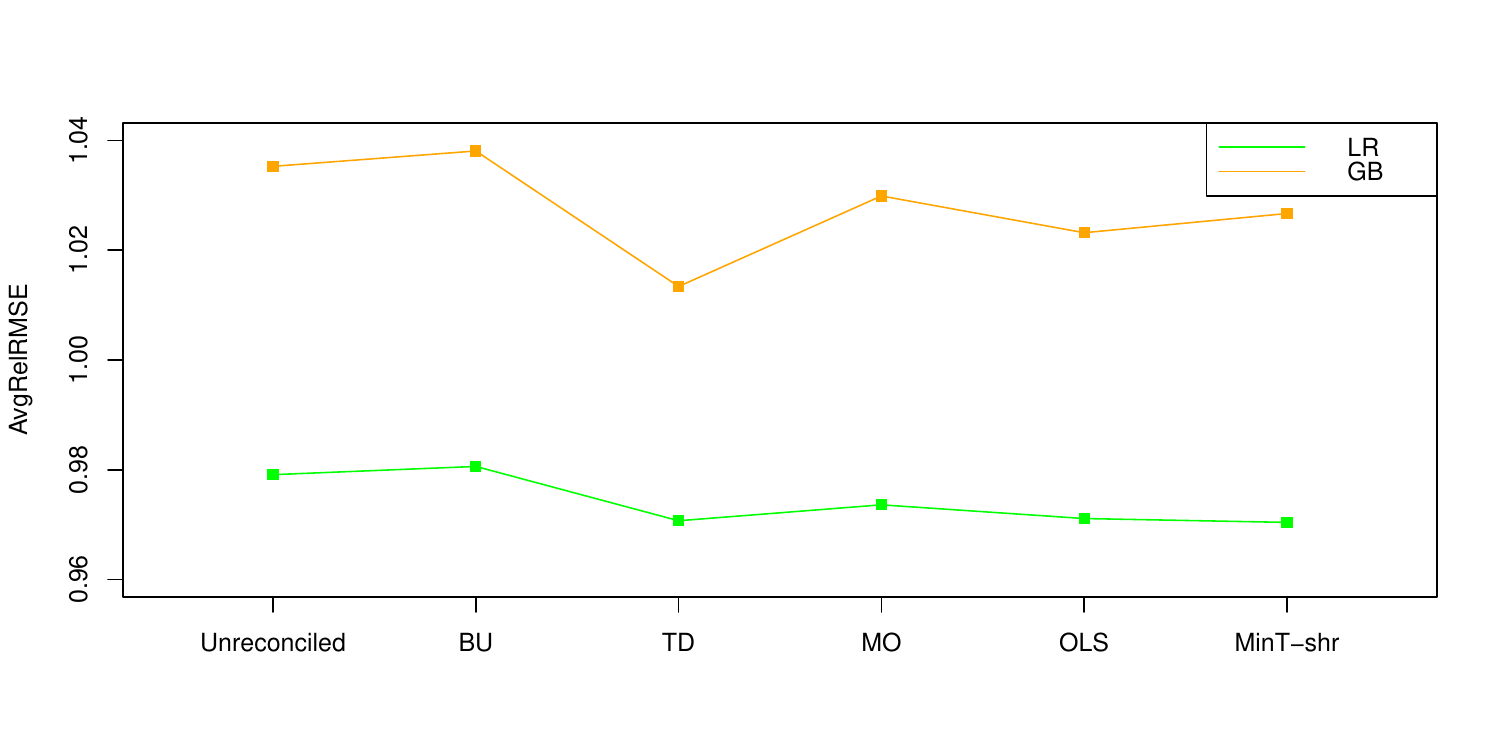}\\
\caption{\textcolor{black}{Forecasting performance of \textcolor{black}{cross-sectional reconciliation} models in the present study across all cross-sectional and temporal aggregations for Data Set B}}
\label{fig:avgrelrmse_B}
\end{figure}

\begin{figure}[H]
\begin{multicols}{2}
    \includegraphics[width=\linewidth]{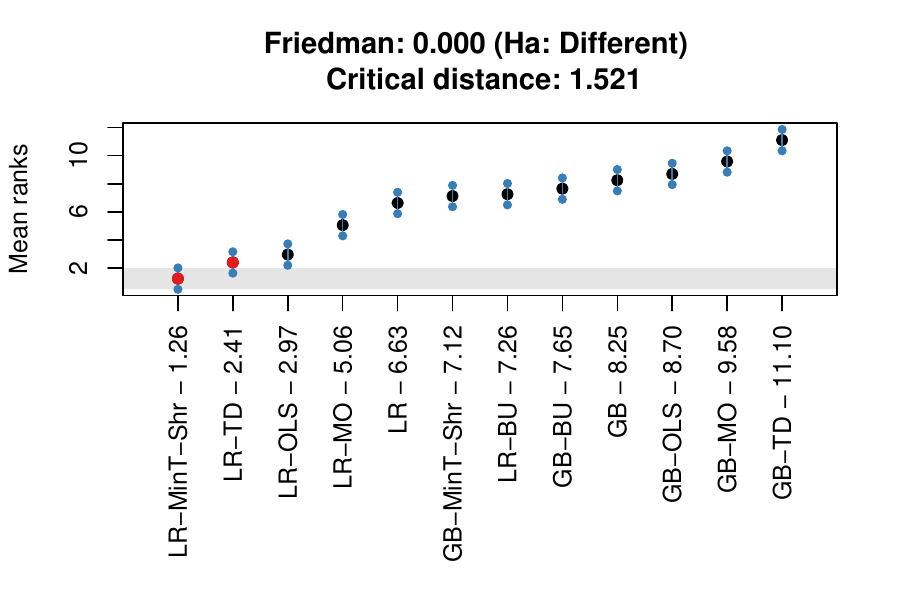}\par
    \includegraphics[width=\linewidth]{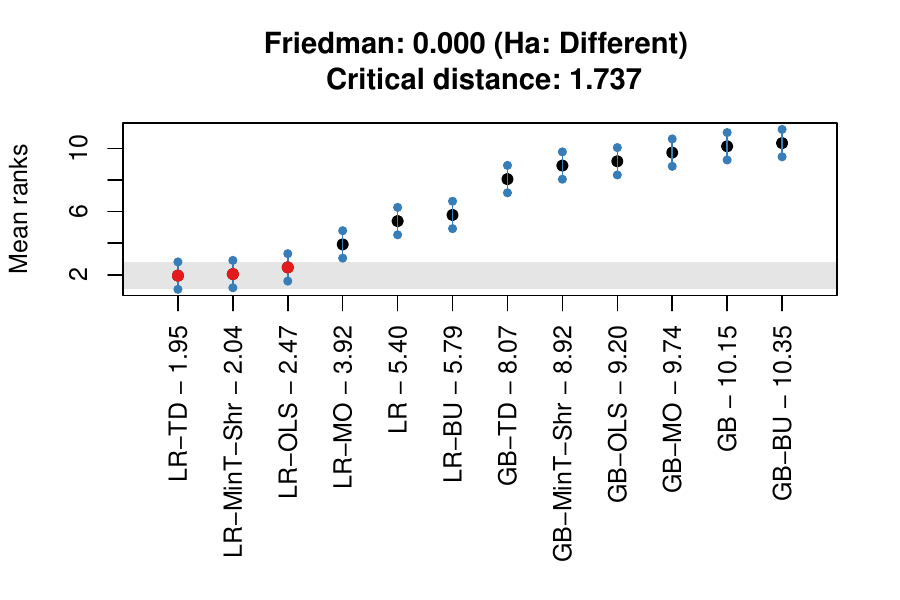}\par
\end{multicols}
\caption{Nemenyi test results at 5\% significance level for all 120 series with both LR and GB base forecasts. The cross-sectional reconciliation methods are sorted horizontally according to the AvgRelRMSE rank for Data Set A (left) and Data Set B (right)}
\label{fig:nemenyi_cross_sec}
\end{figure}

The non-parametric Friedman and post-hoc Nemenyi tests were performed using the \textit{nemenyi()} function from the tsutils package in R. These tests establish if the differences in forecasts produced by the different reconciliation methods are significant \cite{kourentzes2019cross}. The Friedman test establishes whether at least one forecast is significantly different from the rest, and the Nemenyi test identifies groups of forecasts with no evidence of statistically significant differences. The results are shown in Figure \ref{fig:nemenyi_cross_sec}. The Friedman test revealed that there were statistically significant differences in the reconciliation methods for the AvgRelRMSE for LR and GB base forecasts. \textcolor{black}{LR-MinT-Shr} and LR-TD performed best for reconciliation, with no statistical significant difference between them for both data sets. However, for Data Set B, the difference between these methods and LR-OLS was also not statistically significant. This suggests that for cross-sectional reconciliation, linear regression performed better than GB base forecasts, with TD and MinT being the best reconciliation methods. This implies that a simple statistical model such as LR can be used along with appropriate reconciliation techniques for reconciling hierarchical series. Moreover, this can outperform more complicated methods such as GB.

\begin{figure}[H]
\begin{multicols}{2}
    \includegraphics[width=\linewidth]{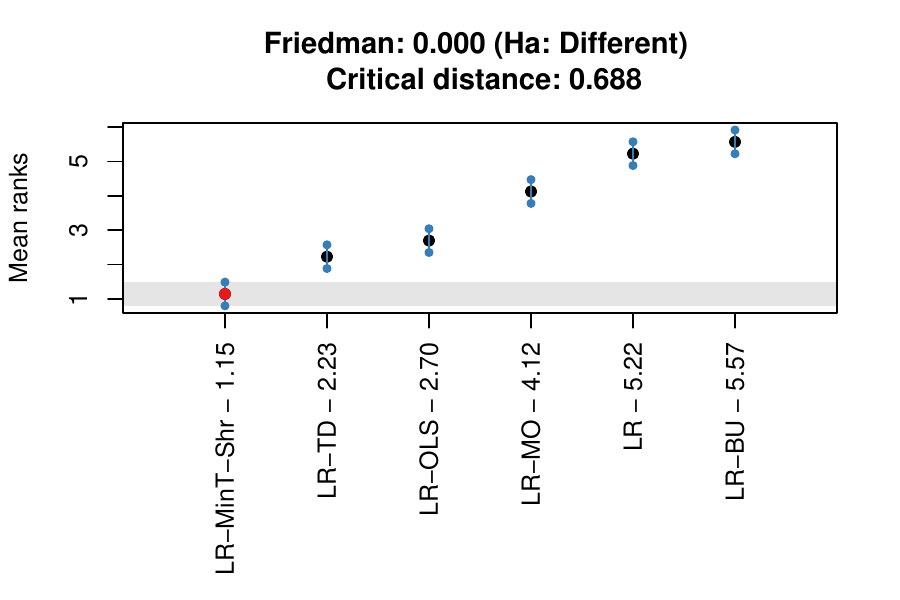}\par
    \includegraphics[width=\linewidth]{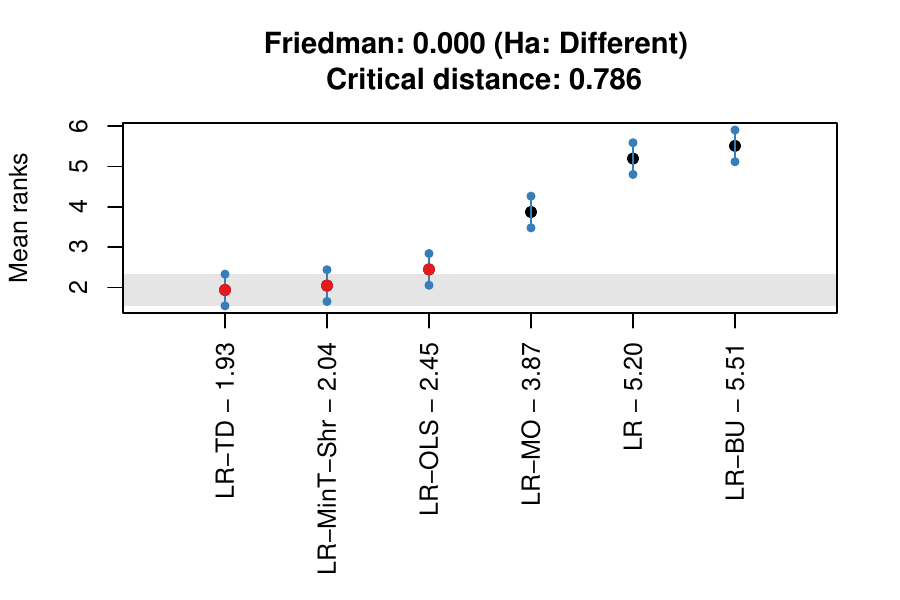}\par
\end{multicols}
\caption{Nemenyi test results at 5\% significance level for all 120 series with only LR base forecasts. The cross-sectional reconciliation methods are sorted horizontally according to the AvgRelRMSE rank for Data Set A (left) and Data Set B (right)}
\label{fig:nemenyi_cross_sec_lr}
\end{figure}

A Nemenyi test was then performed on only the LR base forecasts, and is shown in Figure \ref{fig:nemenyi_cross_sec_lr}. The tests revealed that the AvgRelRMSE for \textcolor{black}{LR-MinT-Shr} was among the best methods for both data sets. However, for Data Set A, all other methods were less accurate and the difference was statistically significant. For Data Set B, LR-TD performed best, and the difference in accuracy with \textcolor{black}{LR-MinT-Shr} and LR-OLS was not statistically significant. The common best method among both data sets was LR-MinT, so this was determined to be the best base forecast and cross-sectional reconciliation method.

\subsection{Cross-temporal reconciliation}\label{cross-temp} 
\vspace{-.35cm}
The results of the AvgRelRMSE, after cross-temporal reconciliation, are given in Tables \ref{tab:results_ct_1}-\ref{tab:results_ct_2}. Values greater than one indicate poorer performance measured through the average relative error index than the naive $h$-steps-ahead benchmarks, \textcolor{black}{and are indicated in red}. Values less than one indicate greater performance. In these tables, LR denotes forecasts produced through linear regression, and GB denotes forecasts produced through gradient boosting. \textcolor{black}{The cross-temporal reconciliation method used is denoted through CT(BU) (cross-temporal bottom-up), T-ACOV (temporal hierarchies using autocovariance), TCS-WLSV-SHR (heuristic first-temporal-then-cross-sectional using temporal series variance and cross-sectional shrinkage), CST-SHR-ACOV (heuristic first-cross-sectional-then-temporal using temporal autocovariance and cross-sectional shrinkage), ITE-ACOV-SHR (iterative cross-temporal reconciliation using temporal series autocovariance and cross-sectional shrinkage) and OCT-WLSV (optimal cross-temporal using series variance scaling).}

\textcolor{black}{At the 10-minutely temporally disaggregated level, for Data Set A, the LR-TCS-WLSV-SHR forecasts outperformed other methods at the bottom level (Level 0), and on average across all levels. Unreconciled LR was the top-performing model at the top level and GB-CT(BU) performed best for aggregated wind turbine forecasts (Level 1). At the 20-minutely level, for Data Set A, LR-TCS-WLSV-SHR was the top-performing model at the bottom level, and across all levels of the cross-sectional hierarchy. For Data Set B, LR-TCS-WLSV-SHR} performed better across all levels. At the 20-minutely level, the forecasting accuracy increases between $1\mbox{-}2\%$ compared to the 10-minutely forecasts. We observe \textcolor{black}{T-ACOV} has among the lowest performance on average, meaning that temporal reconciliation with autocovariance scaling for highly seasonal data such as wind data may not be useful to improve the forecast accuracy. 

At the 30-minutely level, for Data Set A, \textcolor{black}{LR-TCS-WLSV-SHR outperformed other methods at the bottom level, and across all levels of the cross-sectional hierarchy. For Data Set B, LR-TCS-WLSV-SHR performed better than the other methods across all levels. Similarly, the best forecast accuracy at the 30-minutely level is roughly $3\%$ improved compared to the 20-minutely forecasts. We observe the \textcolor{black}{CT(BU)} and TCS-WLSV-SHR} reconciliation methods exhibiting greater performance, and improving the forecast accuracy compared with the 30-minutely cross-sectionally reconciled forecasts. 

At the 1-hourly level, for Data Set A, \textcolor{black}{LR-TCS-WLSV-SHR was the top-performing model at the bottom level, and across all levels of the hierarchy. GB-CT(BU) performed best at the middle and top levels of the hierarchy. For Data Set B, LR-TCS-WLSV-SHR was best across all levels of the hierarchy. We note that while CT(BU) reconciliation methods} may not be attractive from a modelling point of view because it is partially cross-temporal, it is desirable from a practical point of view because it implies that it is better to forecast the energy of each turbine to improve the forecast accuracy at the top level (grid level) which is often of interests to decision makers as opposed to forecasting at the top level directly. At this highest temporally aggregated level, despite GB base forecasts being less accurate than LR forecasts, for Data Set A, the GB-BU-CT exhibited significantly increased accuracy, which suggests that GB may have filtered more of the high-frequency noise which did not improve the aggregated forecasts. In addition, the best reconciliation improved forecast accuracy by roughly $9\mbox{-}10\%$ over 30-minutely forecasts, which is a much larger increase in accuracy compared to the other temporal levels. \textcolor{black}{Nevertheless, we observe a greater performance of cross-temporal forecasts when the base forecasts generated by GB were relatively poor, indicating that cross-temporal reconciliation can significantly improve accuracy for high-frequency data even when base forecasts are worse than naive.} 

\begin{table}[H]\footnotesize
\color{black}
\begin{center}
\caption{Forecasting performance of \textcolor{black}{cross-temporal reconciliation} models in the present study at individual temporal aggregations}\label{tab:results_ct_1}%
\centerline{\begin{tabular}{@{}ll|llll|llll@{}}
\toprule
Data Set & HF Method         & L2 & L1 & L0 & GM & L2 & L1 & L0 & GM\\
\midrule
        &        &  \multicolumn{4}{c|}{\textbf{10-Minutely}} & \multicolumn{4}{c}{\textbf{20-Minutely}}\\
\midrule
A        & LR              & 0.957          & 0.962          & \textbf{0.964} & 0.958 & 0.973            & 0.977        & 0.977 & 0.973 \\
         & LR-CT(BU)       & 0.957          & 0.967          & 0.986 & 0.960 & 0.947            & 0.950        & 0.971 & 0.948 \\
         & LR-T-ACOV$^{*}$       & 0.980          & 0.987          & 0.984 & 0.981 & 0.971            & 0.970        & 0.969 & 0.971 \\
         & LR-TCS-WLSV-SHR & \textbf{0.951}          & 0.963          & 0.972 & \textbf{0.953} & \textbf{0.939}            & 0.945        & 0.957 & \textbf{0.941} \\
         & LR-CST-SHR-ACOV & 0.960          & 0.974          & 0.989 & 0.963 & 0.949            & 0.957        & 0.975 & 0.951 \\
         & LR-ITE-ACOV-SHR & 0.960          & 0.975          & 0.987 & 0.963 & 0.950            & 0.958        & 0.973 & 0.952 \\
         & LR-OCT-WLSV     & 0.965          & 0.982          & \textcolor{red}{1.001} & 0.969 & 0.955            & 0.965        & 0.988 & 0.958 \\
         & GB              & 0.961          & 0.958          & 0.965 & 0.961 & 0.975            & 0.976        & 0.981 & 0.975 \\
         & GB-CT(BU)       & 0.960          & \textbf{0.956}          & 0.970 & 0.960 & 0.947            & \textbf{0.938}        & \textbf{0.955} & 0.946 \\
         & GB-T-ACOV$^{*}$      & 0.980          & 0.984          & 0.989 & 0.981 & 0.968            & 0.965        & 0.971 & 0.968 \\
         & GB-TCS-WLSV-SHR & 0.958          & 0.962          & 0.974 & 0.959 & 0.944            & 0.942        & 0.958 & 0.944 \\
         & GB-CST-SHR-ACOV & 0.970          & 0.978          & 0.995 & 0.972 & 0.956            & 0.959        & 0.978 & 0.958 \\
         & GB-ITE-ACOV-SHR & 0.969          & 0.977          & 0.994 & 0.971 & 0.956            & 0.958        & 0.977 & 0.957 \\
         & GB-OCT-WLSV     & 0.965          & 0.971          & 0.988 & 0.966 & 0.952            & 0.953        & 0.974 & 0.953 \\
         \midrule
B        & LR              & 0.975          & \textbf{0.981}          & \textbf{0.987} & 0.976 & 0.977            & 0.976        & 0.979 & 0.977 \\
         & LR-CT(BU)       & 0.975          & 0.994          & \textcolor{red}{1.010} & 0.978 & 0.954            & 0.960        & 0.974 & 0.955 \\
         & LR-T-ACOV$^{*}$       & \textcolor{red}{1.000}          & \textcolor{red}{1.012}          & \textcolor{red}{1.020} & \textcolor{red}{1.002} & 0.982            & 0.978        & 0.984 & 0.981 \\
         & LR-TCS-WLSV-SHR & \textbf{0.972}          & 0.993          & \textcolor{red}{1.003} & \textbf{0.975} & \textbf{0.951}            & \textbf{0.959}        & \textbf{0.966} & \textbf{0.952} \\
         & LR-CST-SHR-ACOV & 0.984          & \textcolor{red}{1.009}          & \textcolor{red}{1.023} & 0.988 & 0.964            & 0.975        & 0.987 & 0.966 \\
         & LR-ITE-ACOV-SHR & 0.984          & \textcolor{red}{1.008}          & \textcolor{red}{1.022} & 0.988 & 0.964            & 0.974        & 0.985 & 0.966 \\
         & LR-OCT-WLSV     & 0.984          & \textcolor{red}{1.009}          & \textcolor{red}{1.026} & 0.988 & 0.964            & 0.976        & 0.990 & 0.966 \\
         & GB              & \textcolor{red}{1.042}          & \textcolor{red}{1.031}          & \textcolor{red}{1.037} & \textcolor{red}{1.041} & \textcolor{red}{1.035}            & \textcolor{red}{1.020}        & \textcolor{red}{1.029} & \textcolor{red}{1.033} \\
         & GB-CT(BU)       & \textcolor{red}{1.042}          & \textcolor{red}{1.044}          & \textcolor{red}{1.107} & \textcolor{red}{1.045} & \textcolor{red}{1.023}            & \textcolor{red}{1.011}        & \textcolor{red}{1.072} & \textcolor{red}{1.024} \\
         & GB-T-ACOV$^{*}$      & \textcolor{red}{1.053}          & \textcolor{red}{1.052}          & \textcolor{red}{1.059} & \textcolor{red}{1.053} & \textcolor{red}{1.035}            & \textcolor{red}{1.018}        & \textcolor{red}{1.022} & \textcolor{red}{1.033} \\
         & GB-TCS-WLSV-SHR & \textcolor{red}{1.026}          & \textcolor{red}{1.037}          & \textcolor{red}{1.076} & \textcolor{red}{1.029} & \textcolor{red}{1.006}            & \textcolor{red}{1.004}        & \textcolor{red}{1.041} & \textcolor{red}{1.008} \\
         & GB-CST-SHR-ACOV & \textcolor{red}{1.036}          & \textcolor{red}{1.052}          & \textcolor{red}{1.090} & \textcolor{red}{1.040} & \textcolor{red}{1.017}            & \textcolor{red}{1.019}        & \textcolor{red}{1.055} & \textcolor{red}{1.019} \\
         & GB-ITE-ACOV-SHR & \textcolor{red}{1.036}          & \textcolor{red}{1.051}          & \textcolor{red}{1.089} & \textcolor{red}{1.039} & \textcolor{red}{1.017}            & \textcolor{red}{1.018}        & \textcolor{red}{1.053} & \textcolor{red}{1.018} \\
         & GB-OCT-WLSV     & \textcolor{red}{1.037}          & \textcolor{red}{1.049}          & \textcolor{red}{1.100} & \textcolor{red}{1.040} & \textcolor{red}{1.018}            & \textcolor{red}{1.016}        & \textcolor{red}{1.065} & \textcolor{red}{1.020} \\
         \midrule
        &        &  \multicolumn{4}{c|}{\textbf{30-Minutely}} & \multicolumn{4}{c}{\textbf{1-Hourly}}        \\
        \midrule
  A        & LR              & 0.973          & 0.974          & 0.962 & 0.973 & 0.963            & 0.960        & 0.941 & 0.962 \\
         & LR-CT(BU)       & 0.916          & 0.909          & 0.921 & 0.915 & 0.822            & 0.806        & 0.784 & 0.819 \\
         & LR-T-ACOV$^{*}$       & 0.942          & 0.931          & 0.920 & 0.939 & 0.850            & 0.828        & 0.782 & 0.845 \\
         & LR-TCS-WLSV-SHR & \textbf{0.908}          & 0.904          & 0.906 & \textbf{0.907} & \textbf{0.814}            & 0.800        & 0.768 & \textbf{0.810} \\
         & LR-CST-SHR-ACOV & 0.919          & 0.918          & 0.925 & 0.919 & 0.825            & 0.814        & 0.788 & 0.823 \\
         & LR-ITE-ACOV-SHR & 0.919          & 0.918          & 0.923 & 0.919 & 0.826            & 0.815        & 0.785 & 0.823 \\
         & LR-OCT-WLSV     & 0.925          & 0.925          & 0.938 & 0.925 & 0.832            & 0.822        & 0.801 & 0.830 \\
         & GB              & 0.977          & 0.976          & 0.979 & 0.977 & 0.976            & 0.968        & 0.969 & 0.975 \\
         & GB-CT(BU)       & 0.915          & \textbf{0.897}          & \textbf{0.905} & 0.912 & 0.819            & \textbf{0.792}        & \textbf{0.768} & 0.813 \\
         & GB-T-ACOV$^{*}$      & 0.938          & 0.924          & 0.921 & 0.936 & 0.845            & 0.822        & 0.782 & 0.840 \\
         & GB-TCS-WLSV-SHR & 0.911          & 0.901          & 0.907 & 0.910 & 0.816            & 0.796        & 0.770 & 0.812 \\
         & GB-CST-SHR-ACOV & 0.925          & 0.919          & 0.929 & 0.925 & 0.831            & 0.815        & 0.791 & 0.827 \\
         & GB-ITE-ACOV-SHR & 0.925          & 0.918          & 0.927 & 0.924 & 0.830            & 0.814        & 0.790 & 0.827 \\
         & GB-OCT-WLSV     & 0.921          & 0.913          & 0.924 & 0.920 & 0.827            & 0.810        & 0.788 & 0.823 \\
         \midrule
B        & LR              & 0.980          & 0.980          & 0.979 & 0.980 & 0.984            & 0.983        & 0.981 & 0.984 \\
         & LR-CT(BU)       & 0.924          & 0.920          & 0.929 & 0.924 & 0.842            & 0.833        & 0.814 & 0.840 \\
         & LR-T-ACOV$^{*}$       & 0.952          & 0.939          & 0.939 & 0.951 & 0.873            & 0.853        & 0.825 & 0.869 \\
         & LR-TCS-WLSV-SHR & \textbf{0.921}          & \textbf{0.920}          & \textbf{0.921} & \textbf{0.921} & \textbf{0.838}            & \textbf{0.832}        & \textbf{0.807} & \textbf{0.836} \\
         & LR-CST-SHR-ACOV & 0.934          & 0.936          & 0.942 & 0.935 & 0.853            & 0.850        & 0.828 & 0.852 \\
         & LR-ITE-ACOV-SHR & 0.934          & 0.935          & 0.940 & 0.934 & 0.853            & 0.849        & 0.826 & 0.851 \\
         & LR-OCT-WLSV     & 0.934          & 0.937          & 0.945 & 0.935 & 0.854            & 0.851        & 0.832 & 0.852 \\
         & GB              & \textcolor{red}{1.032}          & \textcolor{red}{1.020}          & \textcolor{red}{1.016} & \textcolor{red}{1.030} & \textcolor{red}{1.042}            & \textcolor{red}{1.016}        & \textcolor{red}{1.005} & \textcolor{red}{1.038} \\
         & GB-CT(BU)       & 0.994          & 0.971          & \textcolor{red}{1.027} & 0.993 & 0.913            & 0.885        & 0.913 & 0.910 \\
         & GB-T-ACOV$^{*}$      & \textcolor{red}{1.007}          & 0.978          & 0.976 & \textcolor{red}{1.003} & 0.929            & 0.889        & 0.863 & 0.922 \\
         & GB-TCS-WLSV-SHR & 0.976          & 0.965          & 0.995 & 0.976 & 0.896            & 0.879        & 0.880 & 0.893 \\
         & GB-CST-SHR-ACOV & 0.988          & 0.980          & \textcolor{red}{1.009} & 0.988 & 0.907            & 0.894        & 0.894 & 0.906 \\
         & GB-ITE-ACOV-SHR & 0.987          & 0.979          & \textcolor{red}{1.007} & 0.987 & 0.907            & 0.893        & 0.892 & 0.905 \\
         & GB-OCT-WLSV     & 0.989          & 0.977          & \textcolor{red}{1.020} & 0.989 & 0.909            & 0.892        & 0.906 & 0.908 \\
\bottomrule
\end{tabular}}
\footnotesize{Best solutions in each category are indicated in bold\\
$$^{*}$$ T-ACOV only ensures temporal coherency}
\end{center}
\end{table}

Finally, considering all temporal aggregations in Table \ref{tab:results_ct_2}, for Data Set A, GB-CT(BU) was \textcolor{black}{the top-performing model at Levels 1 and 0 of the cross-sectional hierarchy. LR-TCS-WLSV-SHR was the top performing model at the bottom level and across all levels of the cross-sectional hierarchy. For Data Set B, LR-TCS-WLSV-SHR outperformed other methods across all levels. One dominant trend was the strong performance of the TCS-WLSV-SHR and CT(BU) methods. The strong performance of CT(BU) suggests that there is value in forecasting the wind energy at the lowest level and aggregating them to higher levels which are often of interest to decision-makers, i.e., the power delivered to grid. The strong performance of LR-TCS-WLSV-SHR suggests that addressing temporal dependencies before cross-sectional relationships indicates that capturing and reconciling time-based patterns and trends is crucial for such forecasts. Moreover, in such sequential reconciliation, the cross-sectional shrinkage reconciliation which was found to perform best in Section \ref{cross-sec} is similarly reflected in these results.} For Data Set A, the best cross-temporally reconciled forecasts considering all cross-sectional and temporal levels \textcolor{black}{were $5.1\%$ more} accurate than the best cross-sectionally reconciled forecasts. For Data Set B, the best cross-temporally reconciled forecasts \textcolor{black}{were also $5.1\%$ more} accurate than the best cross-sectionally reconciled forecasts. This suggests that cross-temporal reconciliation exhibits significant advantages over cross-sectional reconciliation, and that for high-frequency wind farm time series, temporally aggregating forecasts at various temporal levels can help to improve the forecast accuracy.

\begin{table}[H]\footnotesize
\color{black}
\begin{center}
\caption{Forecasting performance of \textcolor{black}{cross-temporal reconciliation} models in the present study across all temporal aggregations}\label{tab:results_ct_2}
\centerline{\begin{tabular}{@{}l|llll|llll@{}}
\toprule
         & \multicolumn{8}{c}{\textbf{All temporal aggregations}}                \\
         \midrule
         & \multicolumn{4}{c|}{Data Set A} & \multicolumn{4}{c}{Data Set B} \\
         \midrule
HF Method                 & L2 & L1 & L0 & GM & L2 & L1 & L0 & GM\\
\midrule
LR              & 0.966 & 0.968 & 0.961 & 0.966 & 0.979 & 0.980 & 0.982 & 0.979 \\
LR-CT(BU)       & 0.909 & 0.906 & 0.912 & 0.909 & 0.922 & 0.925 & 0.929 & 0.923 \\
LR-T-ACOV$^{*}$       & 0.934 & 0.927 & 0.910 & 0.932 & 0.951 & 0.944 & 0.939 & 0.949 \\
LR-TCS-WLSV-SHR & \textbf{0.901} & 0.901 & 0.897 & \textbf{0.901} & \textbf{0.919} & \textbf{0.924} & \textbf{0.921 }& \textbf{0.919} \\
LR-CST-SHR-ACOV & 0.912 & 0.914 & 0.915 & 0.912 & 0.933 & 0.941 & 0.942 & 0.934 \\
LR-ITE-ACOV-SHR & 0.912 & 0.914 & 0.913 & 0.912 & 0.932 & 0.940 & 0.940 & 0.933 \\
LR-OCT-WLSV     & 0.918 & 0.921 & 0.928 & 0.919 & 0.933 & 0.941 & 0.945 & 0.934 \\
GB              & 0.972 & 0.969 & 0.973 & 0.972 & \textcolor{red}{1.037} & \textcolor{red}{1.022} & \textcolor{red}{1.022} & \textcolor{red}{1.035} \\
GB-CT(BU)       & 0.909 & \textbf{0.893} & \textbf{0.896} & 0.906 & 0.992 & 0.976 & \textcolor{red}{1.027} & 0.992 \\
GB-T-ACOV$^{*}$      & 0.931 & 0.921 & 0.912 & 0.929 & \textcolor{red}{1.005} & 0.982 & 0.977 & \textcolor{red}{1.002} \\
GB-TCS-WLSV-SHR & 0.905 & 0.898 & 0.898 & 0.904 & 0.975 & 0.969 & 0.995 & 0.975 \\
GB-CST-SHR-ACOV & 0.919 & 0.916 & 0.920 & 0.919 & 0.986 & 0.985 & \textcolor{red}{1.009} & 0.987 \\
GB-ITE-ACOV-SHR & 0.918 & 0.915 & 0.918 & 0.918 & 0.985 & 0.983 & \textcolor{red}{1.008} & 0.986 \\
GB-OCT-WLSV     & 0.914 & 0.910 & 0.915 & 0.914 & 0.987 & 0.982 & \textcolor{red}{1.020} & 0.988\\
\bottomrule
\end{tabular}}
\footnotesize{Best solutions in each category are indicated in bold\\
$$^{*}$$ T-ACOV only ensures temporal coherency}
\end{center}
\end{table}

\begin{figure}[H]
\centering
\includegraphics[scale=0.6]{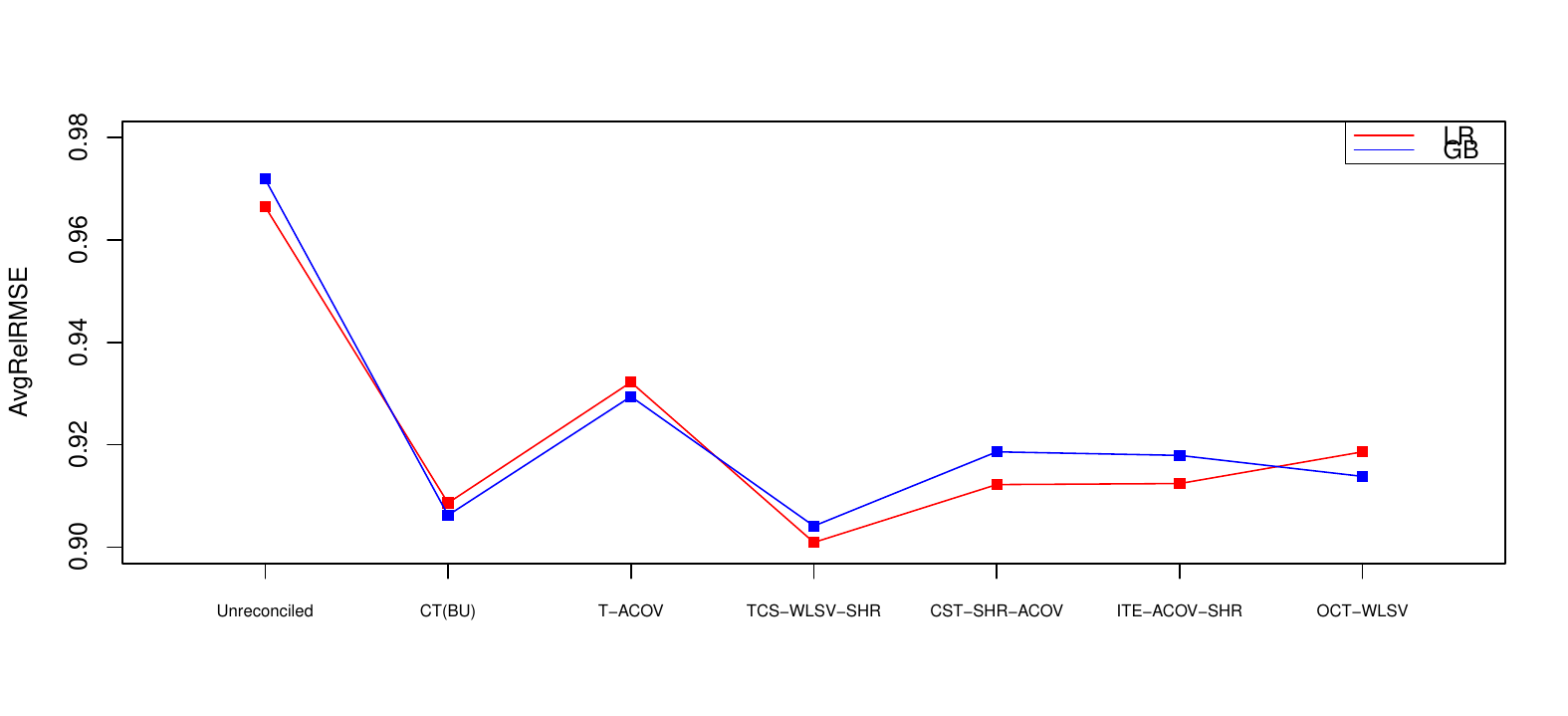}
\caption{\textcolor{black}{Forecasting performance of \textcolor{black}{cross-temporal reconciliation} models in the present study across all cross-sectional and temporal aggregations for Data Set A}}
\label{fig:avgrelrmse_temp}
\end{figure}

\begin{figure}[H]
\centering
\includegraphics[scale=0.6]{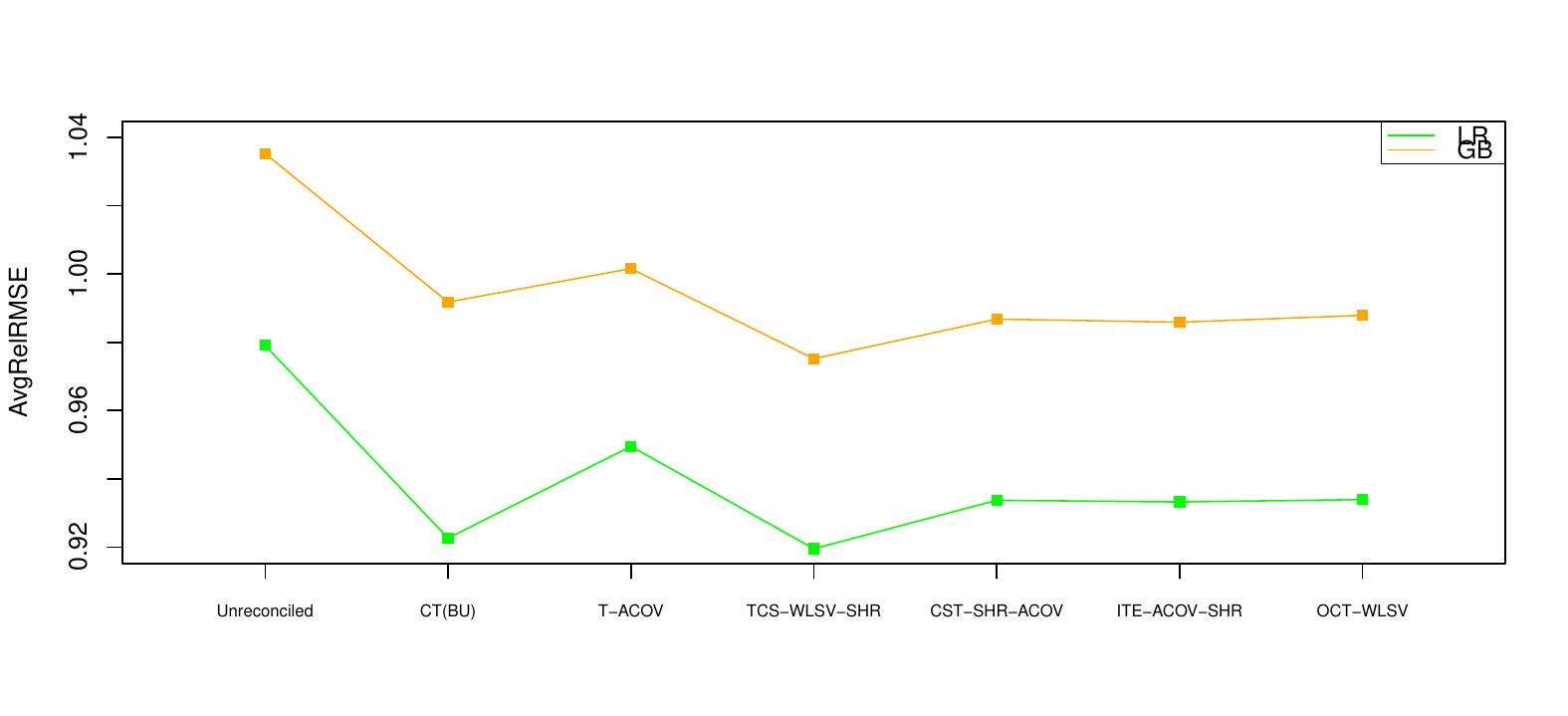}
\caption{\textcolor{black}{Forecasting performance of \textcolor{black}{cross-temporal reconciliation} models in the present study across all cross-sectional and temporal aggregations for Data Set B}}
\label{fig:avgrelrmse_temp2}
\end{figure}

\textcolor{black}{For both data sets, across all cross-sectional and temporal levels, TCS-WLSV-SHR was the best cross-temporal reconciliation method. For Data Set A, LR-TCS-WLSV-SHR} was \textcolor{black}{the top-performing} across all levels. GB produced less accurate base forecasts compared to LR at the 10-minutely level, as shown in Table \ref{tab:results_ct_1}. However, when these base forecasts were aggregated in a bottom-up cross-temporal method, it became the top-performing model at forecasting \textcolor{black}{energy} generated by aggregated turbine models and \textcolor{black}{energy} delivered to the grid. For Data Set B, \textcolor{black}{LR-TCS-WLSV-SHR} was the top-performing across all levels. A comparison of the forecasting performance across all cross-sectional and temporal aggregations is shown in Figures \ref{fig:avgrelrmse_temp}-\ref{fig:avgrelrmse_temp2}. As seen, the performance of LR and GB models are much closer and aligned with each other in comparison to cross-sectional reconciliation results for Data Set A. However, for Data Set B, GB performed significantly worse than LR base forecasts, suggesting greater variability in the applicability of the model to wind farm time series forecasts. For Data Set B, the relative accuracies of the reconciliation methods were very similar.

\begin{figure}[H]
\begin{multicols}{2}
    \includegraphics[width=\linewidth]{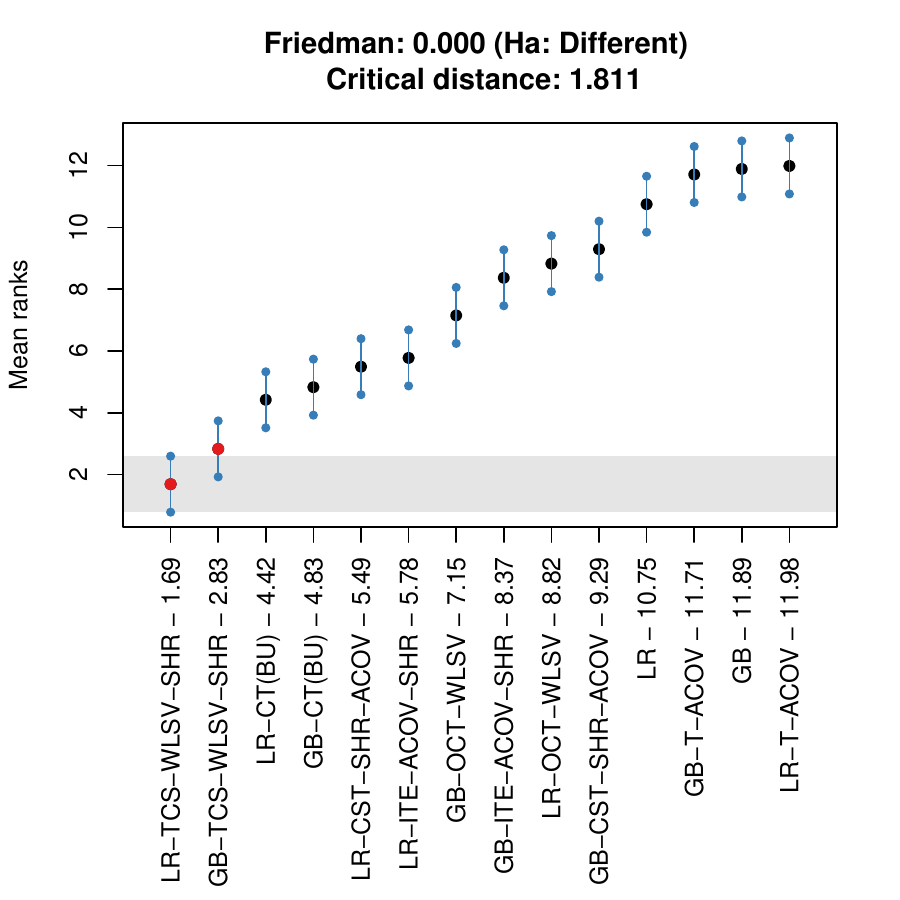}\par
    \includegraphics[width=\linewidth]{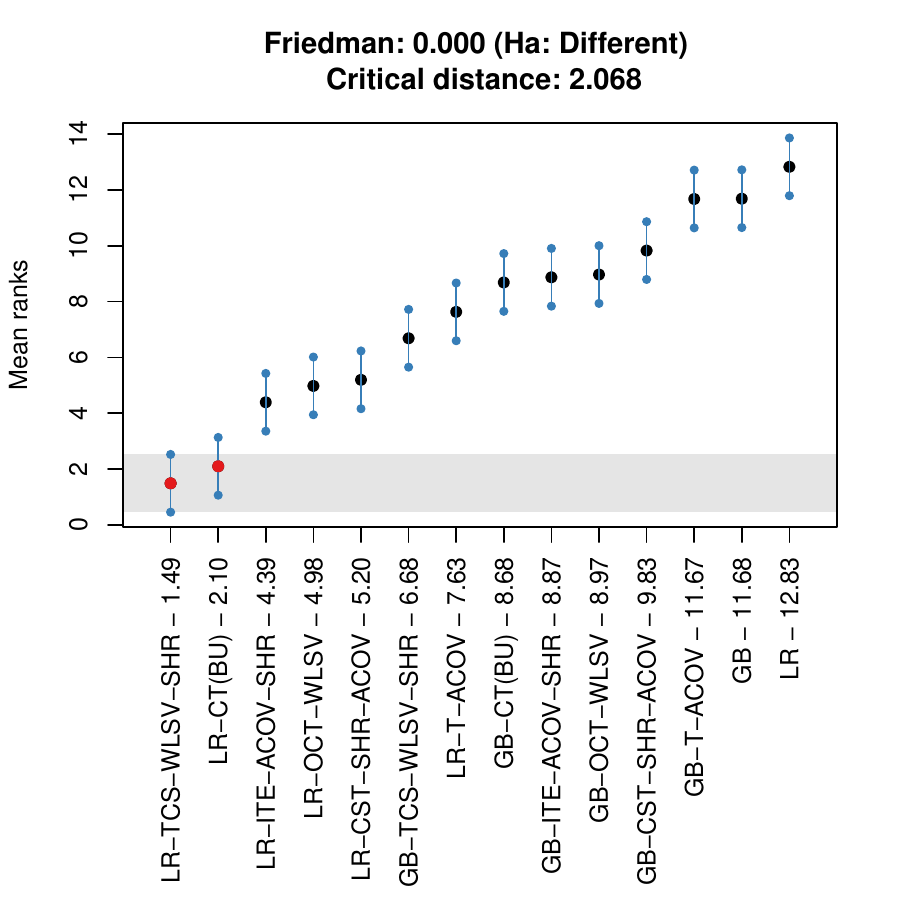}\par
\end{multicols}
\caption{Nemenyi test results at 5\% significance level for all series with both LR and GB  forecasts. The cross-temporal reconciliation methods are sorted horizontally according to the AvgRelRMSE rank for Data Set A (left) and Data Set B (right)}
\label{fig:nem_cross_temp}
\end{figure}

The non-parametric Friedman and post-hoc Nemenyi tests were performed on the cross-temporally reconciled forecasts. The results are shown in Figure \ref{fig:nem_cross_temp} for the AvgRelRMSE. The Friedman test revealed that there were statistically significant differences in the reconciliation methods for AvgRelRMSE with LR and GB base forecasts. For the AvgRelRMSE, \textcolor{black}{LR-TCS-WLSV-SHR and GB-TCS-WLSV-SHR were the best methods for reconciliation, with no statistically significant difference between them for Data Set A. For Data Set B, LR-TCS-WLSV-SHR and LR-CT(BU) forecasts were significantly better than the other methods.}

A Nemenyi test was then performed on only the LR forecasts, which revealed that there was \textcolor{black}{a statistically significant difference between the AvgRelRMSE for the TCS-WLSV-SHR reconciliation method and the others for Data Set A. However, for Data Set B, there was no statistically significant difference between the AvgRelRMSE for the CT(BU) and TCS-WLSV-SHR reconciliation methods.}

\begin{figure}[H]
\begin{multicols}{2}
    \includegraphics[width=\linewidth]{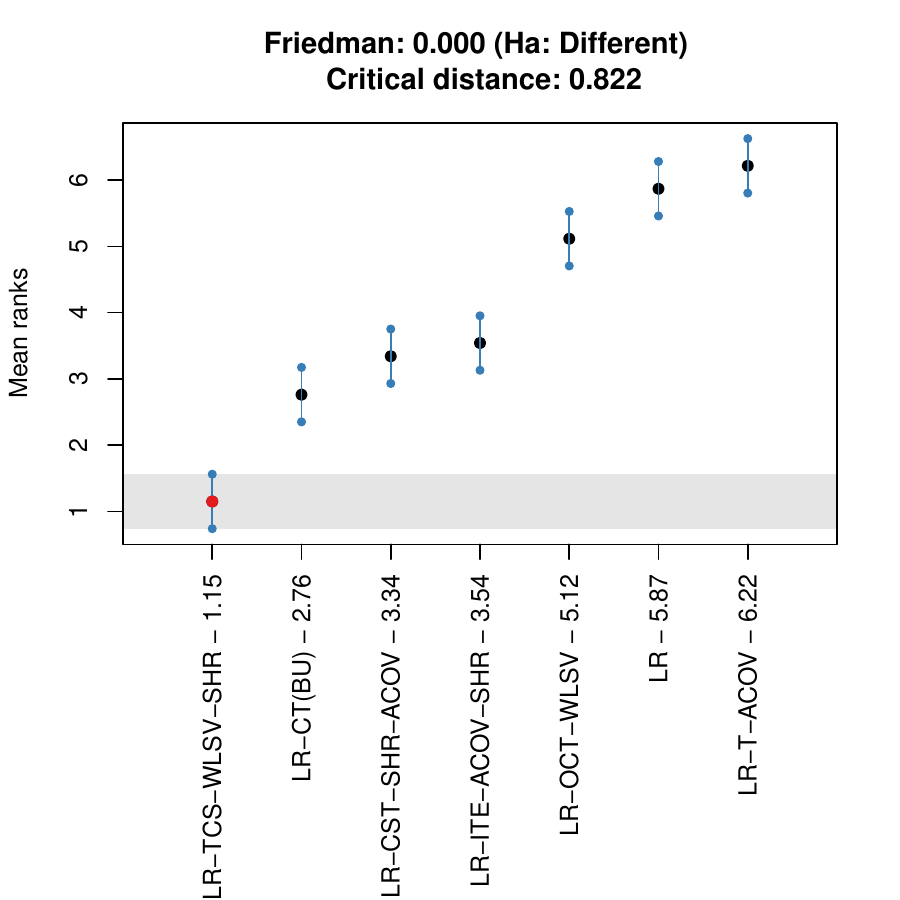}\par
    \includegraphics[width=\linewidth]{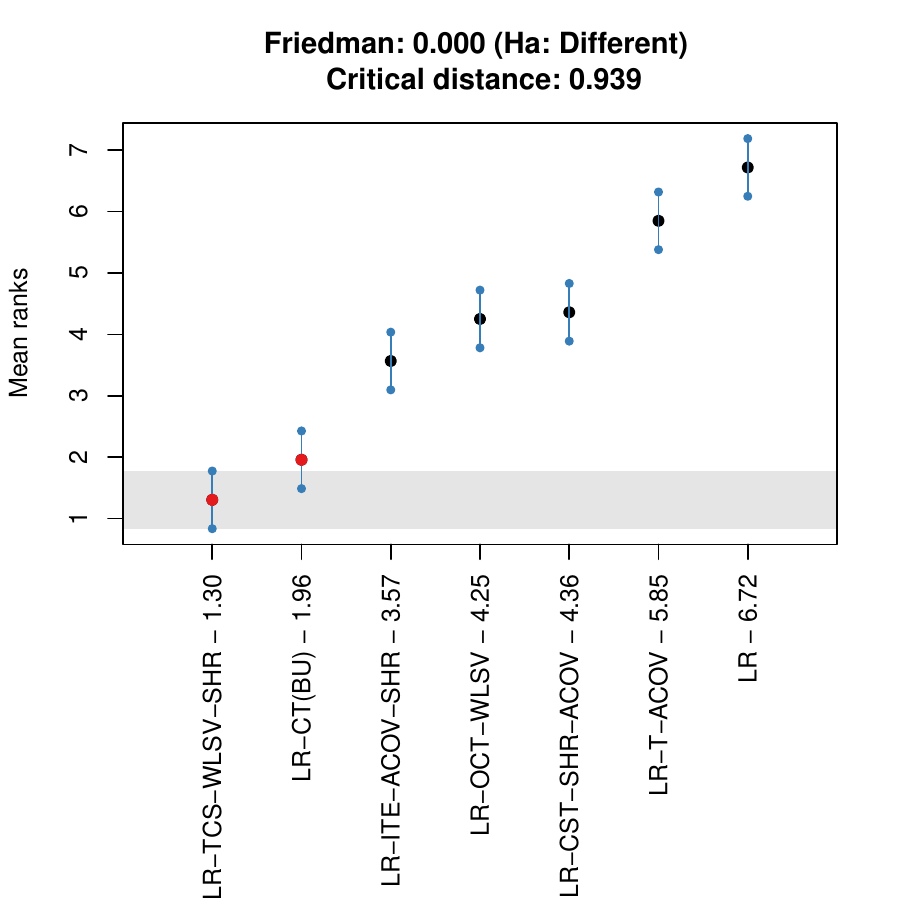}\par
\end{multicols}
\caption{Nemenyi test results at 5\% significance level for all series with only LR  forecasts. The cross-temporal reconciliation methods are sorted horizontally according to the AvgRelRMSE rank for Data Set A (left) and Data Set B (right)}
\label{fig:nem_cross_temp_lr}
\end{figure}

Across all cross-sectional and temporal aggregations, cross-temporal reconciliation significantly improved the forecasting accuracy over cross-sectional reconciliation. The best cross-sectional reconciliation was \textcolor{black}{LR-MinT-Shr which had an AvgRelRMSE of $0.952$ for Data Set A and $0.970$ for Data Set B, whereas the best cross-temporal reconciliation was \textcolor{black}{LR-TCS-WLSV-SHR which had an AvgRelRMSE of $0.901$ for Data Set A and $0.919$ for Data Set B}. This corresponds to a $5.1\%$ (Data Set A) and $5.1\%$ (Data Set B)} increase in forecasting accuracy across all aggregations. However, the best methods at each cross-sectional aggregation level differed. 

\textcolor{black}{In the cross-temporally reconciled paradigm, \textcolor{black}{forecasts across all levels of the cross-sectional hierarchy is better when using LR-TCS-WLSV-SHR for Data Set A. However, this performance relative to GB-CT(BU) decreases up the cross-sectional hierarchy. LR-CT(BU) also performed well in the cross-temporal paradigm.} The base forecasts produced by LR were more accurate than GB, however, bottom-up cross-temporal reconciliation of the GB base forecasts was better at higher levels in the cross-sectional hierarchy. For Data Set B, we found that LR-TCS-WLSV-SHR performed best across all levels in the cross-temporal paradigm. These results indicate that for high frequency wind farm time series, simple base forecasting methods such as LR, combined with aggregating forecasts can outperform more complicated base forecasting and reconciliation methods. This may be partly due to the noise in high-frequency data, but needs further investigation. Sequential temporal-then-cross-sectional reconciliation demonstrated strong forecasting accuracy in this empirical work. Nonetheless, GB shows potential for base forecasts, however the performance at certain cross-sectional levels could not be replicated similarly for both data sets in this work. Hence, one needs to consider the characteristics of the data and the level of forecasts needed, in selecting the best forecasting and reconciliation method.}

\section{Conclusions}\label{sec:conclusion}
This study sought to determine the \textcolor{black}{value of cross-sectional and cross-temporal reconciliation in improving the} forecast accuracy for various horizons and levels in wind farms. In line with the previous studies that had introduced cross-sectional and cross-temporal hierarchies to further improve upon the forecast accuracy \textcolor{black}{of time series, we investigated their value for high-frequency data, i.e., 10-minutely.} Since wind \textcolor{black}{energy} forecasts are often needed across different time horizons, we sought to determine the best practices for generating the most accurate forecast across different horizons in wind farm \textcolor{black}{energy} generation time series, including the level of aggregation in data, choice of cross-sectional and cross-temporal reconciliation methods, and the complexity of base forecasting methods \textcolor{black}{including simple linear regression and light gradient boosting machines}. We evaluated the popular methods of hierarchical forecasting used in literature, for which empirical studies have shown various advantages and disadvantages in different scenarios. The cross-sectional reconciliation algorithms we investigated included those in the bottom-up, top-down, middle-out and combination paradigms. The cross-temporal reconciliation algorithms in this study included bottom-up cross-temporal, temporal hierarchies with autocovariance, heuristic first-temporal-then-cross-sectional, heuristic first-cross-sectional-then-temporal, iterative cross-temporal and optimal cross-temporal reconciliation. 

We generated forecasts for up to one-hour-ahead at different cross-sectional and temporal aggregations using two datasets from wind farms in the UK.  We found linear regression using the minimum trace with shrinkage estimator to be the top-performing model for the cross-sectionally reconciled forecasts across all cross-sectional and temporal aggregations. Unreconciled linear regression forecasts at the root node, that is, \textcolor{black}{energy} delivered to the grid performed better than the other methods at the root cross-sectional level. \textcolor{black}{We also found linear regression with the sequential temporal-then-cross-sectional reconciliation method to perform best for cross-temporally reconciled forecasts across all cross-sectional and temporal aggregations.} Friedman tests indicated that the improvements to forecast accuracies were statistically significant, and post-hoc Nemenyi tests confirmed the best hierarchical forecasting methods.

The cross-temporally reconciled forecasts offer an \textcolor{black}{approximate $5\%$ increase} to accuracy across all cross-sectional and temporal aggregations compared to the cross-sectional reconciliation algorithms. For forecasts that require multiple temporal aggregations for decision-making, cross-temporal reconciliation greatly improves the accuracy compared with simple cross-sectional reconciliation performed on multiple temporal aggregations independently. Moreover, the \textcolor{black}{temporal-then-cross-sectional} reconciliation algorithm was found \textcolor{black}{the top-performing model in our experiment}. \textcolor{black}{However, bottom-up cross-temporal forecasts also performed well,} which requires only base forecasts at the finest granularity, reducing computational forecasting costs. In addition, this work showed that \textcolor{black}{gradient boosting methods may be a good choice for generating base forecasts for high-frequency data but simple models like linear regression offer greater accuracy in most of the scenarios.}

Future research could investigate the value of new reconciliation methods tailored for wind farms, for instance by including the spatial correlation and distance between turbines. \textcolor{black}{It is possible to customise the loss functions of the forecasting models to incorporate desired features into the learning process \cite{abolghasemi2021effectively}.} Various computational techniques and experiment setups such as optimizing hyperparameters, and implementing different base forecasting methods such as generalized logit transformations of the wind \textcolor{black}{energy} time series to better model the sigmoid-resembling power curves, especially for probabilistic very short-term wind \textcolor{black}{energy} forecasts, could also be considered. It would be useful to evaluate the methodologies in this study in other wind farms to investigate whether the results found in this work are reproducible for other wind farms with different turbines and structures. \textcolor{black}{Finally, as only one year of data was used for both data sets in this work, future research should consider whether the empirical results found in this work are applicable to longer data sets, which may exhibit slow-changing trends or time-dependent seasonalities.}


\bibliographystyle{unsrturl}  
\bibliography{references} 

\newpage

\appendix

\section{Descriptive statistics of wind \textcolor{black}{energy}}
We conducted an exploratory data analysis for each temporally disaggregated time series (i.e., 10-minutely) in the cross-sectional hierarchy for both data sets. The minimum value, first quartile, median, mean, third quartile and maximum values are tabulated in Tables \ref{tab:descriptive} and \ref{tab:descriptiveY}.

From Table \ref{tab:descriptive}, for Data Set A, turbine model B delivered the greatest \textcolor{black}{energy} to the grid, followed by model D, model C and lastly model A. Observe that model D has fewer turbines than model A, but each turbine generates more \textcolor{black}{energy} on average than those of model A. All turbines had a minimum value of 0.0 kWh, which represented a 10-minutely period during which the turbine did not generate any \textcolor{black}{energy}. Models A-C also shared (approximately) a greatest maximum value of approximately $460$ kWh, which was the upper limit for the turbines. For turbines of model D, this upper limit was instead around $506$ kWh.

\begin{table}[h!]\footnotesize
\color{black}
\begin{center}
\caption{Descriptive statistics of \textcolor{black}{energy} generated by wind turbines for Data Set A}\label{tab:descriptive}%
\centerline{\begin{tabular}{llllllll}
\toprule
Model & Turbine & Min & 1st Quartile & Median & Mean & 3rd Quartile & Max \\
\midrule
A     & A1      & 0.0 & 45.8         & 126.4  & 171.2  & 288.9        & 460.3  \\
A     & A2      & 0.0 & 35.0         & 107.1  & 157.4  & 265.4        & 460.6  \\
A     & A3      & 0.0 & 41.7         & 121.7  & 168.2  & 287.5        & 460.2  \\
A     & A4      & 0.0 & 38.7         & 116.7  & 165.5  & 282.4        & 460.2  \\
A     & A5      & 0.0 & 42.1         & 121.3  & 164.8  & 277.3        & 460.2  \\
A     & A6      & 0.0 & 43.4         & 123.1  & 168.7  & 283.3        & 460.2  \\
A     & A7      & 0.0 & 39.9         & 119.2  & 166.1  & 283.0        & 460.4  \\
A     & Agg.    & 4.8 & 306.1        & 868.1  & 1161.9 & 1957.5       & 3220.8 \\
\midrule
B     & B1      & 0.0 & 33.9         & 103.2  & 151.9  & 255.0        & 460.3  \\
B     & B2      & 0.0 & 36.5         & 110.6  & 156.9  & 268.0        & 460.3  \\
B     & B3      & 0.0 & 38.0         & 114.2  & 154.0  & 260.2        & 460.2  \\
B     & B4      & 0.0 & 29.5         & 96.8   & 144.8  & 246.4        & 460.3  \\
B     & B5      & 0.0 & 16.7         & 83.8   & 139.7  & 239.4        & 460.2  \\
B     & B6      & 0.0 & 34.9         & 100.1  & 145.6  & 236.9        & 460.2  \\
B     & B7      & 0.0 & 28.7         & 88.2   & 135.7  & 224.8        & 460.2  \\
B     & B8      & 0.0 & 32.5         & 97.9   & 146.7  & 247.3        & 460.1  \\
B     & B9      & 0.0 & 39.7         & 121.4  & 164.2  & 284.1        & 460.2  \\
B     & Agg.    & 9.5 & 332.8        & 953.5  & 1339.5 & 2263.2       & 4140.9 \\
\midrule
C     & C1      & 0.0 & 38.8         & 120.3  & 163.0  & 276.4        & 460.2  \\
C     & C2      & 0.0 & 29.0         & 105.4  & 152.3  & 261.7        & 460.2  \\
C     & C3      & 0.0 & 35.8         & 104.8  & 150.2  & 247.5        & 460.2  \\
C     & Agg.    & 0.6 & 114.9        & 334.9  & 465.5  & 778.0        & 1380.4 \\
\midrule
D     & D1      & 0.0 & 36.4         & 125.7  & 185.8  & 323.1        & 506.3  \\
D     & D2      & 0.0 & 45.7         & 142.9  & 200.9  & 349.1        & 506.3  \\
D     & D3      & 0.0 & 48.1         & 155.4  & 209.1  & 375.8        & 506.3  \\
D     & D4      & 0.0 & 53.8         & 148.0  & 193.3  & 316.8        & 506.3  \\
D     & D5      & 0.0 & 43.8         & 147.7  & 204.3  & 366.3        & 506.3  \\
D     & D6      & 0.0 & 53.7         & 161.7  & 212.7  & 377.3        & 506.3  \\
D     & Agg.    & 3.8 & 321.1        & 899.8  & 1206.1 & 2089.9       & 3037.2 \\
\bottomrule
\end{tabular}}
\end{center}
\color{black}
\end{table}

From Table \ref{tab:descriptiveY}, for Data Set B, model A delivered greater \textcolor{black}{energy} to the grid than model B since it included more turbines. However, the turbines of both models generated approximately the same average \textcolor{black}{energy}, and had equal maximum \textcolor{black}{energy} limits. The average \textcolor{black}{energy} generated was smaller than the turbines of Data Set A, due to differences in turbine models. In addition, the cross-sectional hierarchy structures were different, with Data Set A having a balanced structure, while Data Set B only included two models with most turbines being of a single model. However, the inter-turbine variance is similar, which allows for a reasonable comparison between the two sets of data.

\begin{table}[H]\footnotesize
\color{black}
\begin{center}
\caption{Descriptive statistics of \textcolor{black}{energy} generated by wind turbines for Data Set B}\label{tab:descriptiveY}%
\centerline{\begin{tabular}{llllllll}
\toprule
Model & Turbine & Min & 1st Quartile & Median & Mean & 3rd Quartile & Max \\
\midrule
A     & A1      & 0.0 & 18.3         & 85.1   & 127.5  & 216.8        & 346.1  \\
A     & A2      & 0.0 & 14.5         & 74.5   & 120.5  & 205.0        & 347.6  \\
A     & A4      & 0.0 & 17.0         & 82.4   & 125.7  & 212.3        & 347.2  \\
A     & A5      & 0.0 & 14.2         & 72.6   & 118.6  & 203.4        & 345.5  \\
A     & A6      & 0.0 & 14.5         & 72.7   & 119.7  & 204.4        & 345.8  \\
A     & A7      & 0.0 & 10.1         & 59.9   & 108.7  & 199.4        & 345.0  \\
A     & A8      & 0.0 & 14.9         & 73.0   & 117.4  & 208.1        & 345.4  \\
A     & A9      & 0.0 & 10.2         & 56.4   & 103.6  & 187.7        & 345.2  \\
A     & A10     & 0.0 & 8.8          & 52.3   & 101.8  & 190.4        & 345.6  \\
A     & A11     & 0.0 & 16.6         & 73.7   & 119.9  & 204.5        & 346.0  \\
A     & A12     & 0.0 & 22.4         & 90.7   & 130.6  & 221.1        & 345.2  \\
A     & A13     & 0.0 & 21.8         & 88.4   & 129.7  & 218.3        & 346.5  \\
A     & A14     & 0.0 & 20.6         & 85.1   & 128.7  & 219.8        & 345.9  \\
A     & A15     & 0.0 & 22.5         & 86.5   & 129.7  & 219.8        & 345.9  \\
A     & Agg.    & 7.3 & 244.4        & 1109.1 & 1682.2 & 2828.0       & 4803.7 \\
\midrule
B     & B1      & 0.0 & 22.2         & 80.3   & 119.4  & 198.2        & 346.4  \\
B     & B2      & 0.0 & 28.6         & 93.9   & 131.1  & 225.5        & 346.9  \\
B     & B3      & 0.0 & 21.6         & 67.8   & 110.1  & 176.2        & 346.6  \\
B     & B4      & 0.0 & 23.2         & 82.0   & 120.5  & 199.7        & 347.1  \\
B     & B5      & 0.0 & 21.6         & 72.3   & 114.7  & 188.3        & 347.0  \\
B     & B6      & 0.0 & 16.9         & 59.7   & 101.0  & 155.8        & 345.9  \\
B     & Agg.    & 1.8 & 152.3        & 474.1  & 696.7  & 1138.3       & 2059.5 \\
\bottomrule
\end{tabular}}
\end{center}
\color{black}
\end{table}

\end{document}